\documentclass[preprint,12pt,authoryear,a4paper]{elsarticle}
\usepackage[utf8]{inputenc}
\usepackage[T1]{fontenc}
\usepackage{amsmath,amssymb,mathtools,amsthm}
\usepackage{graphicx,booktabs,multirow,hhline,makecell}
\usepackage[table]{xcolor}
\usepackage{microtype,nicefrac,url,pifont,wrapfig,subcaption,enumitem}
\usepackage[export]{adjustbox}
\usepackage{algorithm,algpseudocode}
\usepackage[textsize=tiny]{todonotes}
\usepackage{hyperref}
\hypersetup{hidelinks}
\biboptions{round,semicolon}
\journal{ISPRS Journal of Photogrammetry and Remote Sensing}

\definecolor{mygray}{gray}{.91}
\newlength\savewidth

\newcommand{\TODOtext}[1]{\textcolor{red}{\textbf{[\,TODO\ifx\relax#1\relax\else:~#1\fi\,]}}}
\newcommand{\TODO}[1]{\ifmmode\mbox{\TODOtext{#1}}\else\TODOtext{#1}\fi}

\usepackage[section]{placeins}
\usepackage{flafter}
\usepackage{float}
\newenvironment{evidencewithtext}{%
  \par\pagebreak[0]\edef\savedfigureparindent{\the\parindent}%
  \noindent\begin{minipage}{\linewidth}%
  \setlength{\parindent}{\savedfigureparindent}\ignorespaces
}{\end{minipage}\par\addvspace{\intextsep}}


\begin{document}
\begin{frontmatter}
    \title{High-fidelity training-free photovoltaic module mapping from multi-view UAV imagery}
    \author[cuggie]{WenHao~Duan}
    \author[cuggie]{Qi~Chen\corref{cor1}}
    \ead{chenqi@cug.edu.cn} 
    \author[eitndt,eitlab]{Penglei~Zhang}
    \author[eitndt,eitlab,polyu]{Ruizhe~Deng}
    \author[polyu]{Zhiling~Guo}
    \author[pku]{Haoran~Zhang}
    \author[eitndt,eitlab]{Yuntian~Chen}
    \cortext[cor1]{Corresponding author.}
    \affiliation[cuggie]{organization={School of Geography and Information Engineering, China University of Geosciences (Wuhan)},citysep={},city={Wuhan},postcode={430074},country={China}}
    \affiliation[eitndt]{organization={Ningbo Institute of Digital Twin, Eastern Institute of Technology},city={Ningbo},statesep={},state={Zhejiang},postcode={315200},country={China}}
    \affiliation[eitlab]{organization={Zhejiang Key Laboratory of Industrial Intelligence and Digital Twin, Eastern Institute of Technology},city={Ningbo},statesep={},state={Zhejiang},postcode={315200},country={China}}
    \affiliation[polyu]{organization={Department of Building Environment and Energy Engineering, The Hong Kong Polytechnic University},addressline={Kowloon},city={Hong Kong},country={China}}
    \affiliation[pku]{organization={School of Urban Planning and Design, Peking University},city={Shenzhen},postcode={518055},country={China}}

    \begin{abstract}
        High-fidelity photovoltaic (PV) module mapping requires complete instance recovery and accurate delineation. Vision foundation models offer zero-shot segmentation, but converting incomplete and inconsistent predictions across views into a complete, geometrically accurate map of individual modules remains challenging. We propose the Multi-View PV Module Mapper (MV-PVMapper), a training-free framework coupling vision foundation models with unmanned aerial vehicle (UAV) photogrammetry. First, Segment Anything Model~3 (SAM~3) generates candidates from original images using text prompts. Their contours are then transferred into object space through per-module slope correction to limit distortion from surface-model noise, and verified against representative module dimensions to guide cross-view instance selection. Finally, accepted instances return to all covering images as box prompts for SAM~3, supplementing text prompts with spatial cues for iterative segmentation, verification and fusion. On five-direction surveys of three operational rooftop PV plants at approximately 1\,cm nadir ground sampling distance, two feedback iterations recover 7\,710 of 7\,711 reference modules with no false positives at an intersection-over-union (IoU) matching threshold of 0.5. Macro-averaged area precision and recall reach 99.00\% and 98.52\%, respectively. Averaged across sites, the aggregated Jaccard index reaches 97.55\%, compared with 60.73\% and 41.24\% for orthomosaic segmentation and multi-view pixel-voting baselines, respectively; average precision at the stricter IoU threshold of 0.95 reaches 95.67\%. Ablation studies show that module-size constraints are critical for instance separation, while slope correction and feedback to all covering views improve boundary accuracy. These results are achieved without task-specific image annotation or segmentation-model weight updates.
    \end{abstract}
\end{frontmatter}

\section{Introduction}
\label{sec:intro}

The global energy transition is rapidly expanding the installed base of photovoltaic (PV) systems. By the end of 2025, global cumulative PV capacity had approached 3\,TW, after an estimated 698\,GW was installed in that year alone\citep{De_lEpine2026-wg}. As the installed base grows, maintaining reliable performance over the lifetime of these assets requires inspection and maintenance at an increasingly large scale\citep{GALLARDOSAAVEDRA2018566,CHEN2023120579,barraz2025towards}.

Within large PV installations, tasks such as targeted diagnosis, repair and replacement require inspection findings to be linked to individual modules and their locations. Module-level georeferencing supports this link between image-based inspection and field operations\citep{barraz2025towards,bommes2022georeferencing}. A module map provides a common spatial representation in which each module has its own boundary and location. Its quality depends on recovering the module inventory while preserving the separation and geometry of individual instances, rather than only identifying the area occupied by PV arrays. Unmanned aerial vehicle (UAV) surveys can cover an entire plant efficiently, but converting UAV images into such a map remains difficult. Modules are densely and repeatedly arranged, their gaps may span only a few pixels, and their apparent scale and shape change under oblique viewing.

PV module mapping involves both extracting individual instances and establishing their geographic locations. Rule-based methods can avoid labelled data, but depend on assumptions about colour, line structure, shape or array layout\citep{wang2018photovoltaic,cardoso2024automated}. Supervised detectors and segmentation networks can learn more flexible appearance features, but require annotated UAV images. Even domain-generalization methods rely on training data collected across different module types and imaging conditions to improve performance on unseen sites\citep{barraz2024one}. Beyond recognising and delineating modules, mapping also requires their image-space detections to be placed in a common geographic frame. Photogrammetry provides this spatial reference for UAV imagery. Structure-from-motion (SfM) estimates camera poses and reconstructs scene geometry, supporting both the georeferencing of modules detected in individual images and the generation of orthomosaics. These products support different mapping workflows: modules can be extracted from a georeferenced digital orthophoto map (DOM)\citep{cardoso2024automated}, or observations in multiple images can be associated and located using the reconstructed geometry\citep{bommes2022georeferencing,kozak2025detailed}.

Promptable visual foundation models (VFMs) offer a way to reduce the annotation burden. The Segment Anything Model (SAM) supports zero-shot transfer to new image distributions and tasks through prompts\citep{kirillov2023segment}. SAM~3 further accepts concept prompts and can detect and segment instances named by a short noun phrase\citep{carion2025sam}. Recent work has shown the potential of this training-free, open-vocabulary approach by formulating remote-sensing image interpretation as two-dimensional segmentation\citep{li2025segearthov3}. Within a photogrammetric workflow, the same model can be applied directly to a georeferenced DOM, so the predicted masks already have ground locations and physical extents. However, segmentation on a DOM uses a single orthorectified representation rather than explicitly combining predictions from the overlapping source images. For densely arranged PV modules, observations from different viewpoints can provide complementary evidence where individual views leave boundaries unclear.

A multi-view UAV oblique survey provides overlapping images that can observe the same module from several directions, while camera poses and a digital surface model (DSM) relate these observations within a common object-space frame. In addition, PV modules within a plant commonly follow one or a few standardized size specifications, allowing representative dimensions to be established from a small number of field measurements. Using these sources jointly for module mapping, however, raises two related challenges. The first is to transfer predictions reliably into object space despite noise in the reconstructed surface, which can distort their locations and shapes. The second is to combine predictions from views with different visibility and image quality, so that complementary observations reinforce one another while conflicting results are suppressed.

This study aims to achieve high-fidelity PV module mapping by combining zero-shot VFM segmentation with redundant observations from UAV photogrammetry, without task-specific training. We propose the Multi-View PV Module Mapper (MV-PVMapper), which links image-space segmentation and object-space geometric verification through iterative multi-view feedback. First, SAM~3 generates module candidates from the original images using a text concept prompt, without site-specific image annotation or segmentation-model weight updates. Second, per-module slope correction transfers candidate contours into object space while limiting deformation caused by DSM noise. Third, segmentation confidence and consistency with representative module dimensions guide cross-view fusion, retaining plausible module instances while suppressing conflicting and duplicate predictions. Finally, accepted object-space instances are projected into covering images as box prompts; the resulting masks are transferred back for geometric verification and fusion, providing updated spatial evidence for subsequent segmentation rounds. We evaluate MV-PVMapper on five-direction oblique photogrammetric data from three operational rooftop PV plants, with a nadir ground sampling distance (GSD) of approximately 1\,cm/pixel. After two feedback iterations, MV-PVMapper matches 7\,710 of the 7\,711 reference modules across all three sites, with no false positives at an intersection-over-union (IoU) threshold of 0.5. It achieves macro-averaged area precision and recall of 99.00\% and 98.52\%, respectively. Its macro-averaged aggregated Jaccard index (AJI) and average precision at the stricter IoU threshold of 0.95 ($AP_{95}$) reach 97.55\% and 95.67\%, compared with 60.73\% and 60.51\% for an orthomosaic segmentation baseline (DOM-IterSeg), and 41.24\% and 27.52\% for a multi-view pixel-voting baseline (MV-PixelVote). Together, these results show near-complete instance recovery and accurate delineation on the evaluated sites, rather than high pixel precision alone.

The main contributions are as follows:
\begin{itemize}
    \item We propose MV-PVMapper, a training-free framework that couples segmentation by a pretrained VFM with multi-view instance fusion in an iterative refinement loop for PV module mapping. By feeding geometrically verified instances back to all covering views as spatial prompts, the framework uses complementary observations to recover missed modules and refine their boundaries.

    \item We introduce a per-module slope-correction strategy for projecting PV masks from image space to object space. By adapting the projection to each module's local surface, it preserves module geometry and improves robustness to DSM noise.

    \item We convert readily available module dimensions into an object-space geometry score that guides candidate selection, cross-view association and feedback. This geometry-guided strategy suppresses over- and under-segmentation and helps limit error accumulation across iterations.
\end{itemize}

\section{Related Work}
\label{sec:related}

We review PV extraction methods, prompting strategies for VFMs, and multi-view fusion with geometric priors. The review distinguishes task-specific learning from training-free inference and examines how spatial evidence can support instance association and delineation, providing the context for MV-PVMapper.

\subsection{PV Module Extraction and Mapping}
\label{sec:rw-pv-detection}

Image-based PV extraction supports applications ranging from installation inventories to component inspection \citep{CHEN2023120579,barraz2025towards}. Traditional methods identify panels through explicit appearance and geometric rules. Association between region and line primitives can support automatic template generation and matching without training samples \citep{wang2018photovoltaic}. Edge detection, morphological processing and contour extraction provide another route to delineating individual panels \citep{cardoso2024automated}. These methods make use of the regular appearance of PV installations, but their applicability depends on the chosen rules and thresholds. Changes in illumination, panel appearance or the visibility of narrow gaps can weaken the cues on which extraction relies.

Supervised learning replaces much of this manual feature design with representations learned from task data. For UAV thermal imagery, training across different module types and acquisition conditions has been used to improve generalisation to unseen installations \citep{barraz2024one}. Other approaches reduce the cost of adapting models to new data: weakly supervised domain adaptation transfers rooftop PV segmentation from densely labelled source imagery to target imagery with sparse point annotations \citep{wang2026epuda}, while parameter-efficient VFM adaptation has been explored for PV change detection \citep{huang2026pvpadapter}. These developments address training-data diversity and annotation cost, but still require task-specific learning. Their practical deployment therefore still involves obtaining suitable training data and adapting the model.

\subsection{Prompting and Refinement with Visual Foundation Models}
\label{sec:rw-sam}

SAM allows points, boxes and masks to specify the object to be segmented \citep{kirillov2023segment}. Evaluations on overhead imagery show useful zero-shot transfer, but also failures associated with the characteristics of remote-sensing targets \citep{ren2024samspace}. One response is to learn task-specific prompts or output representations. Category-aware prompt generation supports automatic instance segmentation \citep{chen2024rsprompter}, while automatic box prompting and a decoder trained to predict vertices and boundaries support polygonal building extraction \citep{wang2024sampolybuild}. In PV imagery, learned multi-scale and edge-pyramid prompts incorporate scale and boundary information into SAM adaptation \citep{yao2026pvsam}. These methods use the foundation model as a basis for task-specific learning.

Language prompts offer a convenient way to specify a target category without placing a spatial prompt on each object. This can be implemented by coupling a text-conditioned detector with SAM \citep{ren2024groundedsam}, or through the unified concept-prompting interface of SAM~3 \citep{carion2025sam}. Direct application of SAM~3 to remote-sensing semantic segmentation has also been explored without additional training \citep{li2025segearthov3}. PV-specific vision--language pretraining offers a complementary approach: PANEL supports downstream zero- and few-shot interpretation after learning from a PV image--text corpus \citep{deng2026panel}. MV-PVMapper instead uses a generic VFM without PV-specific training and coordinates its predictions across UAV views to recover georeferenced module instances.

The prompt interface also supports refinement: previous masks and corrective prompts can be supplied in subsequent segmentation rounds \citep{kirillov2023segment}. The quality of this refinement depends on the evidence supplied to the model. In spatial mapping, observations from other views and geometric information provide possible sources of such evidence, alongside the prediction in the current image.

\subsection{Multi-view Fusion and Geometric Priors}
\label{sec:rw_mvia}

Photogrammetric geometry allows observations from different images to be compared in a common spatial frame. In PV mapping, detections have been georeferenced through tracking and SfM, or associated across images using semantic keypoints and plant structure \citep{bommes2022georeferencing,kozak2025detailed}. Learned fusion also uses this spatial correspondence to combine information before prediction. Image features can be backprojected and pooled on a voxel grid together with geometric features \citep{dai2018_3dmv}, or aggregated on 3D points with weights learned from viewing conditions \citep{robert2022deepviewagg}. For aerial building mapping, MVMapper combines original-image and object-space features using supervised learning and piecewise affine projection to reduce DSM-induced misalignment \citep{chen2024mvmapper}. MV-PVMapper shares MVMapper's use of a common spatial frame to exploit complementary aerial observations, but operates on predicted module instances rather than learned feature fusion. Geometric verification and cross-view feedback coordinate these instances while the segmentation model remains frozen.

At the instance level, fusion must distinguish repeated detections from neighbouring objects. Non-maximum suppression (NMS) ranks candidates and removes those with excessive overlap; Soft-NMS instead reduces their scores according to overlap \citep{bodla2017softnms}. For multi-view masks, the evidence can extend beyond pairwise overlap. Consistency with masks from several views can guide the merging of 3D regions \citep{yin2024sai3d}, and agreement among views can define graph connections for clustering masks into instances \citep{yan2024maskclustering}. Spatial information can also guide segmentation itself: SAMPro3D projects shared 3D prompts into images and selects and consolidates them using multi-view mask evidence, without additional training \citep{xu2025sampro3d}. Thus, geometric information can support both instance association and prompt selection.

Module dimensions and regular layouts provide constraints beyond visual appearance. In PV applications, physical module dimensions have been used to generate training boxes automatically \citep{barraz2025fast}. Detected templates and layout constraints can also guide the completion of module maps \citep{cardoso2024automated}. Such priors can complement VFMs: SinkSAM-Net uses topographic depressions to generate prompts and shape information to filter SAM masks before training a downstream segmenter \citep{rafaeli2025sinksam}. MV-PVMapper uses representative dimensions to verify object-space candidates and select instances for mapping and subsequent spatial prompting.

\section{Datasets}
\label{sec:datasets}

We constructed a UAV oblique photogrammetric dataset of three operational rooftop PV plants. For each plant, it contains image-space data, namely the original nadir and oblique images, and object-space data, namely the camera poses, DSM and DOM reconstructed from these images, together with a reference polygon for every module.

\subsection{Study Sites and UAV Acquisition}
\label{sec:sites}

The three plants are installed on building roofs and contain 7\,711 modules in total (Table~\ref{tab:sites}). Sites~1 and~2 contain 1\,772 and 2\,305 modules, respectively, each mounted at a single height, whereas the 3\,634 modules at Site~3 are mounted at two heights about 10\,m apart.

\begin{table}[!htp]
\centering
\caption{Study sites and UAV acquisition. Module size is the representative length $\times$ width
of the modules at each site, taken in this study as the mean length $\times$ width of the reference
modules.}
\label{tab:sites}
\footnotesize
\begin{tabular}{lccc}
\toprule
 & \textbf{Site 1} & \textbf{Site 2} & \textbf{Site 3} \\
\midrule
Reference modules          & 1\,772             & 2\,305             & 3\,634 \\
Module size (m)            & $2.22 \times 1.09$ & $2.17 \times 1.10$ & $2.20 \times 1.09$ \\
Mounting levels            & 1                  & 1                  & 2 \\
Mounting type              & Roof-flush         & Racked             & Roof-flush \\
\midrule
Survey date                & 2025-10            & 2025-11            & 2024-08 \\
Relative flight height (m) & 35.0               & 38.0               & 41.8 \\
Oblique gimbal pitch       & $-45^{\circ}$      & $-45^{\circ}$      & $-60^{\circ}$ \\
Nadir GSD (cm/pixel)       & 0.93               & 1.01               & 1.11 \\
Images (nadir / oblique)   & 62 / 266           & 94 / 289           & 329 / 1\,125 \\
Coordinate system (EPSG code)       & 4550               & 4550               & 4547 \\
\bottomrule
\end{tabular}
\end{table}

All sites were surveyed with a DJI Mavic~3E, whose wide-angle colour camera records $5280 \times 3956$ pixel images at a focal length of 12.29\,mm. Each survey combined nadir images with oblique images from four directions at 80\% forward and 70\% side overlap, giving 2\,165 images for the three sites. The median flight heights above the take-off point were 35.0 to 41.8\,m, corresponding to a nadir GSD of 0.93 to 1.11\,cm/pixel. The oblique gimbal pitch was $-45^{\circ}$ at Sites~1 and~2 and $-60^{\circ}$ at Site~3. Site~3 was surveyed in August 2024, and Sites~1 and~2 in October and November 2025.

\subsection{Photogrammetric Processing and Reference Generation}
\label{sec:photogrammetry}

\begin{figure}[!htp]
    \centering
    \includegraphics[width=0.9\textwidth]{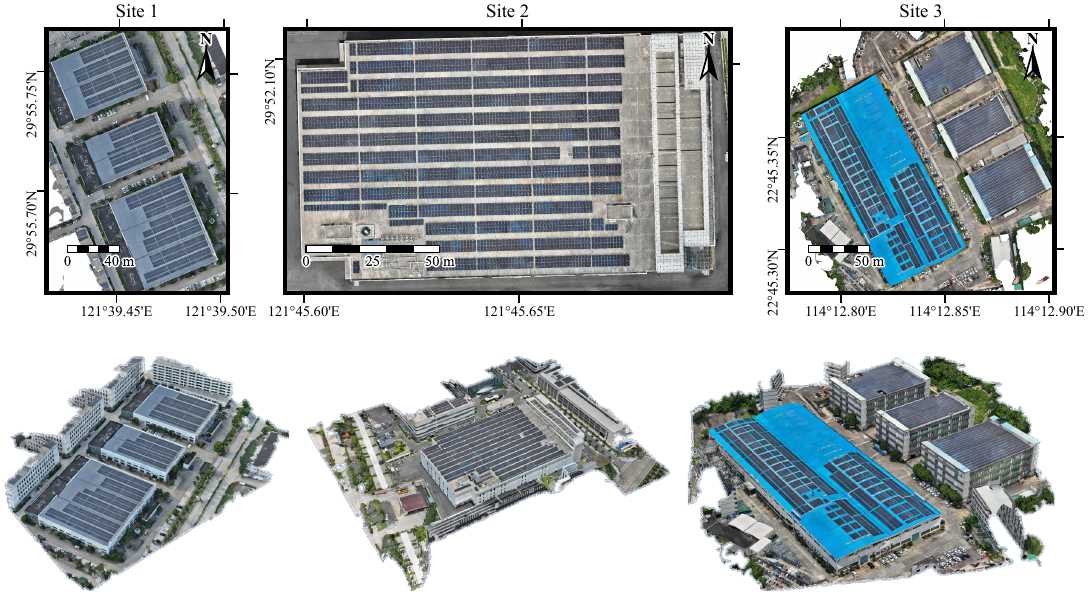}
    \caption{DOMs (top) and oblique views of the reconstructed 3D models (bottom) of Sites~1--3.}
    \label{fig:data_stations}
\end{figure}

The images of each site were processed in ContextCapture. Aerotriangulation of all nadir and oblique images estimated the camera poses, with the image positions recorded by the onboard real-time kinematic receiver providing the georeferencing. Dense matching produced the DSM, from which a true orthophoto, referred to below as the DOM, was generated. Figure~\ref{fig:data_stations} shows the DOM and the reconstructed 3D model of each site. Every module is represented by one reference polygon, giving 7\,711 reference polygons for the three sites.

\section{Methods}
\label{sec:methods}
\begin{figure}[!htp]
    \centering
    \includegraphics[width=\textwidth]{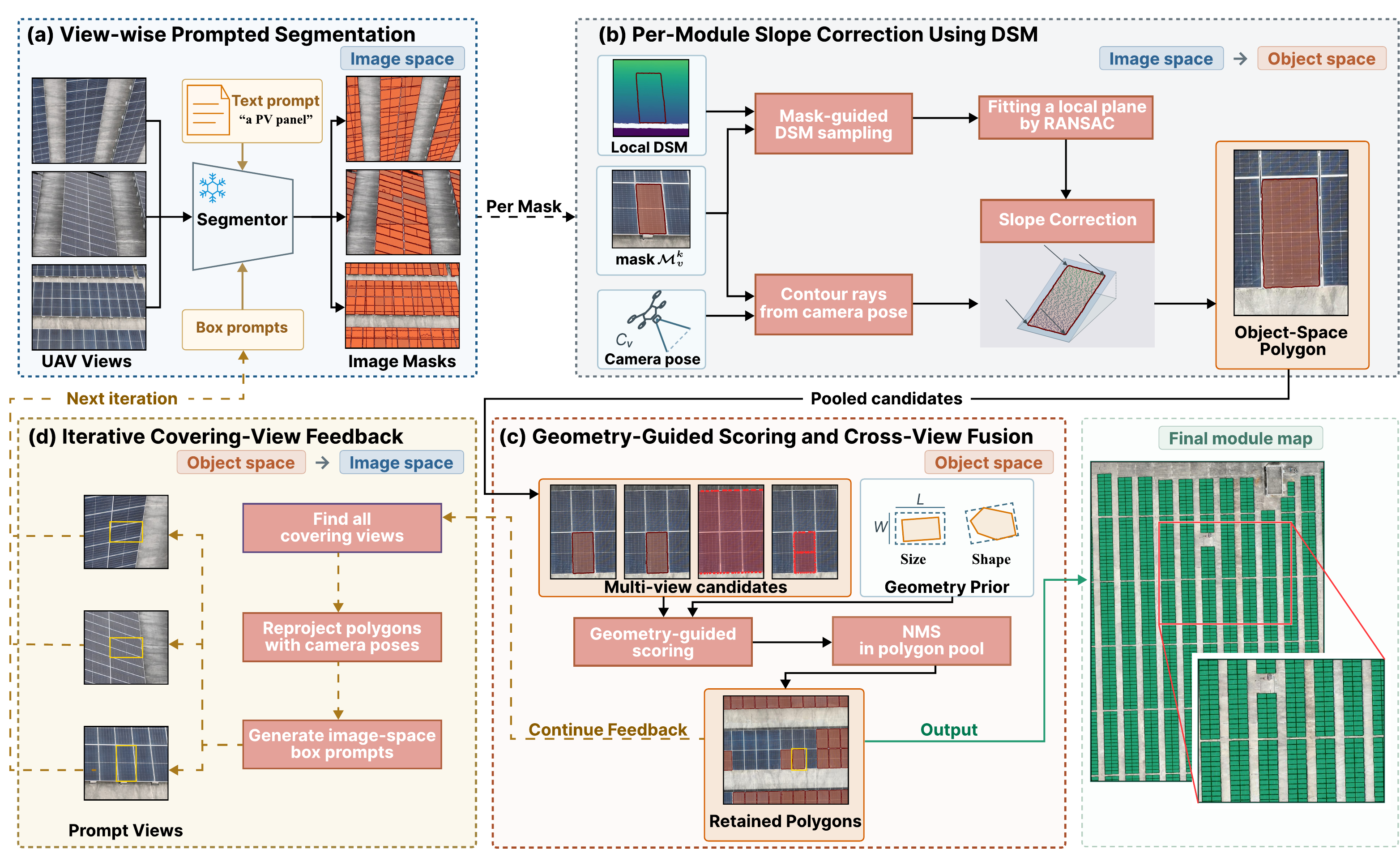}
    \caption{Workflow of MV-PVMapper. Panels (a)--(d) correspond to
    Sections~\ref{sec:sam3}--\ref{sec:iterative-pipeline}. The snowflake marks the frozen weights of the
    segmentor, the promptable segmentation model described in Section~\ref{sec:sam3}. $C_v$ is the
    camera pose of image~$v$, $\mathcal{M}_v^k$ is the mask of its $k$-th candidate, and $L$ and $W$ are
    the representative module length and width.}
    \label{fig:pipeline}
\end{figure}

MV-PVMapper takes overlapping UAV images, calibrated camera poses, a georeferenced DSM and representative module dimensions as inputs, and outputs a map of separate, georeferenced module polygons (Figure~\ref{fig:pipeline}). It alternates between prompted segmentation in image space and geometric verification and fusion in object space. Per-module slope correction transfers candidate contours to a common spatial frame, where segmentation confidence and module geometry guide cross-view instance selection. The accepted instances return to the covering images as box prompts for further segmentation. The final module map is the accepted instance set from the last fusion round.

\subsection{View-Wise Prompted Segmentation}
\label{sec:sam3}

Iterative segmentation requires prompts that can evolve as spatial evidence becomes available. A text prompt identifies the target category and provides initial candidates without known module locations. Once reliable instances have been identified, box prompts add location cues that a repeated category description alone cannot provide. The segmentation model must therefore accept text and box prompts and return instance masks with confidence scores. We use SAM 3, whose concept-prompting and interactive segmentation capabilities support these requirements \citep{carion2025sam}.

For the initial pass, SAM~3 is queried with the text prompt \texttt{a PV panel}. Inference uses overlapping windows to preserve the narrow gaps between neighbouring modules, which may become difficult to resolve if an entire UAV image is resized to the model input. The last row and column of windows are aligned with the image boundary, and the predicted masks are restored to the original image coordinates using the window offsets. Candidates below the confidence or mask-area threshold are discarded.

In subsequent rounds, accepted object-space instances are projected back into the covering images to generate box prompts (Section~\ref{sec:iterative-pipeline}). By conveying module locations established from other views, these box prompts guide the re-segmentation of modules previously missed or poorly delineated in the target image. Each box is clipped to every window it intersects. For each window, all available box prompts are used unless their number exceeds $N_b$, in which case $N_b$ prompts are randomly selected. Windows with box prompts use both spatial and text prompts, whereas windows without box prompts use text-only inference. Duplicates from overlapping windows are retained until object-space fusion, where they are resolved together with cross-view duplicates (Section~\ref{sec:postprocessing}). Numerical settings are given in Section~\ref{sec:configuration}.

\subsection{Per-Module Slope Correction Using DSM}
\label{sec:slopecorrection}

Photogrammetric DSMs can smooth sharp height discontinuities at object edges. Intersecting each contour vertex's viewing ray with such a surface can therefore displace or distort an otherwise accurate image-space contour \citep{chen2024mvmapper}. In dense PV arrays, these errors affect module edges and the narrow gaps between rows.

We introduce a per-module slope-correction strategy that uses the approximately planar module surface to project PV mask contours into object space. For each mask, a z-buffer identifies visible DSM points that fall within the mask when projected into the image. We then fit a local plane to these points using random sample consensus (RANSAC) \citep{fischler1981random}, reducing the influence of outliers near module edges or on surrounding surfaces. Candidates are excluded if the fitted plane is insufficiently supported by the points or has an implausibly steep slope. For each retained candidate, the camera pose and interior orientation define a viewing ray through each mask-contour vertex. We intersect these rays with the fitted plane and use the planimetric coordinates of the intersections to construct the object-space polygon. The contour is thus transferred using a common local plane, rather than independent intersections with the DSM, preserving straight edges without imposing a rectangular shape. Whether the resulting polygon corresponds to one physical module is assessed in Section~\ref{sec:postprocessing}.
\par

Figure~\ref{fig:abl_projection} illustrates the effect of the local surface model. Collinearity-based projection transfers local height errors to the contour vertices and can bend edges across module gaps. A two-dimensional image-to-object affine transformation \citep{chen2024mvmapper} preserves straight edges, but DSM points from the surrounding roof can bias the estimated affine parameters. Slope correction projection instead uses RANSAC to fit the surface from visible DSM points that project into the mask, giving closer agreement with the module boundaries in this example.

\ifdefined\ablprojW\else \newlength{\ablprojW}\setlength{\ablprojW}{0.238\textwidth}\fi
\providecommand{\ablprojdir}{figs_PV/abl_projection}
\providecommand{\ablprojpanel}[1]{\includegraphics[width=\ablprojW]{\ablprojdir/#1.png}}

\begin{figure}[!htp]
    \centering
    \captionsetup[subfigure]{justification=centering}
    \begin{subfigure}[t]{\ablprojW}
        \ablprojpanel{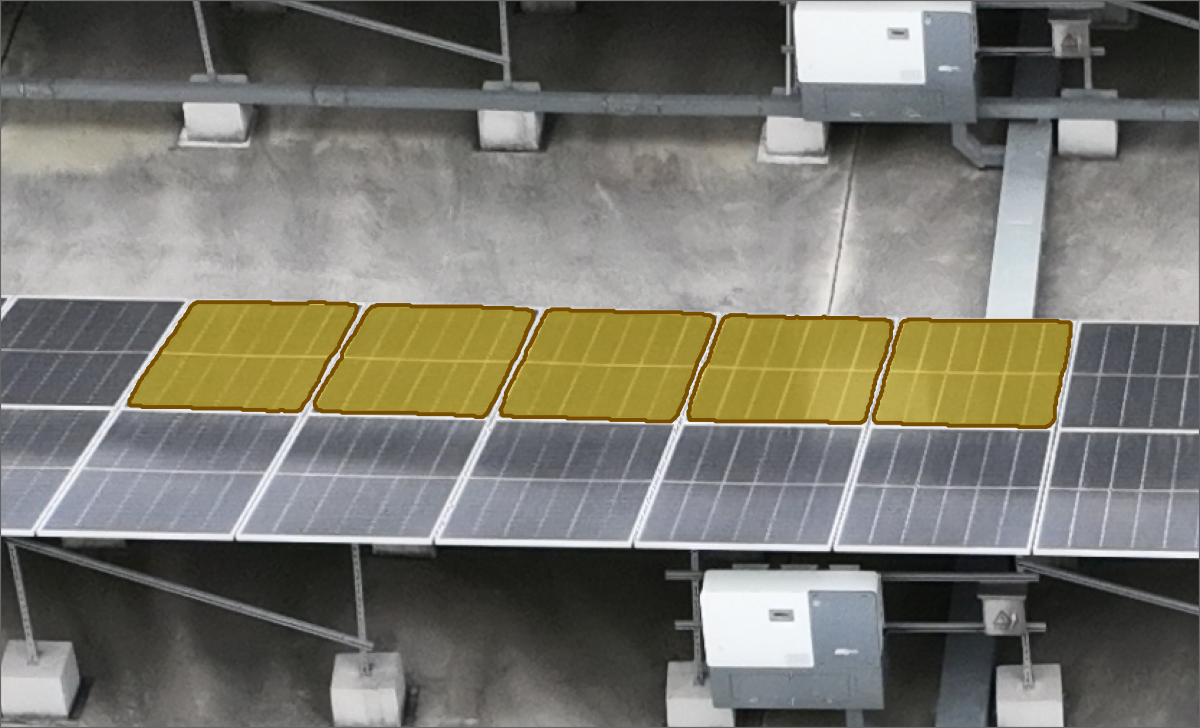}
        \caption{Image space}
        \label{fig:abl_proj_image}
    \end{subfigure}%
    \hfill%
    \begin{subfigure}[t]{\ablprojW}
        \ablprojpanel{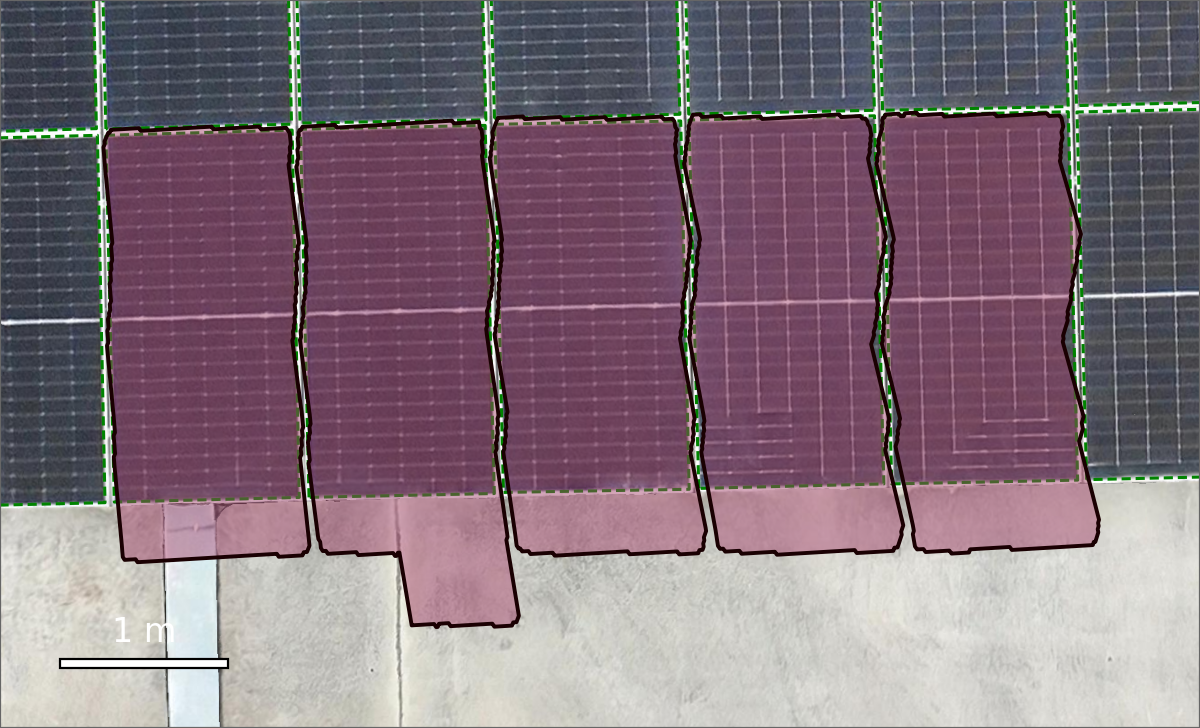}
        \caption{Collinearity-based projection}
        \label{fig:abl_proj_collinearity}
    \end{subfigure}%
    \hfill%
    \begin{subfigure}[t]{\ablprojW}
        \ablprojpanel{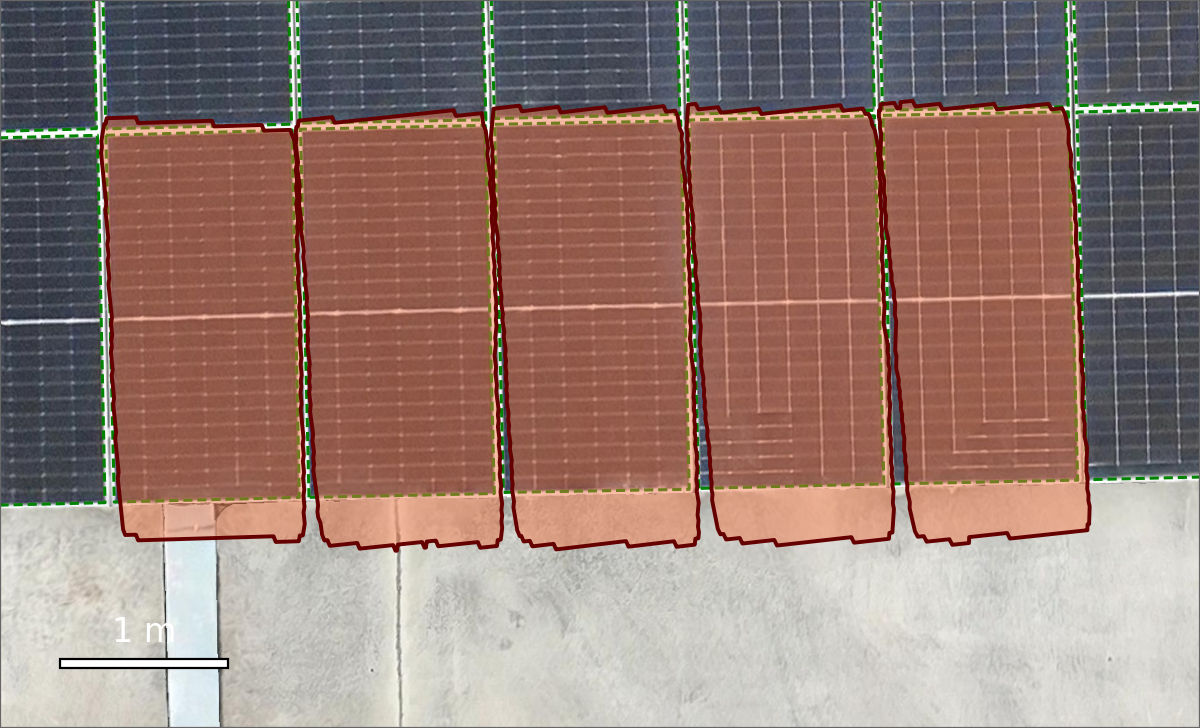}
        \caption{Affine-based projection }
        \label{fig:abl_proj_affine}
    \end{subfigure}%
    \hfill%
    \begin{subfigure}[t]{\ablprojW}
        \ablprojpanel{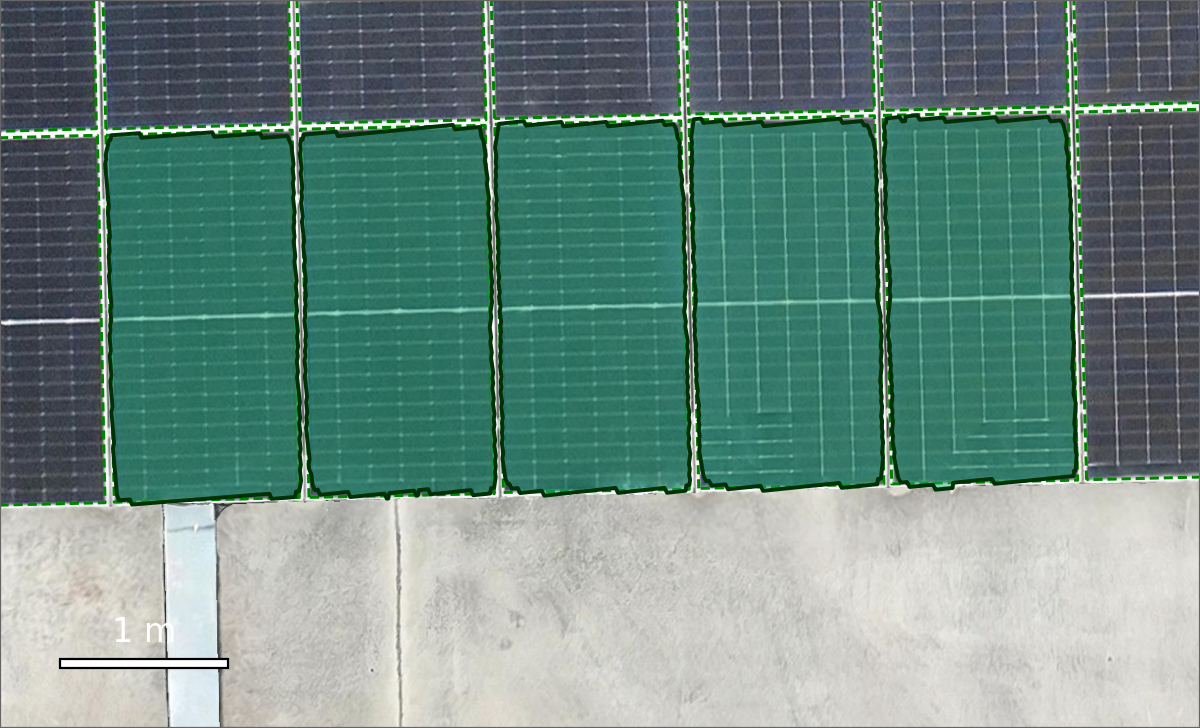}
        \caption{Slope correction projection}
        \label{fig:abl_proj_slope}
    \end{subfigure}
    \caption{Object-space contours of a row of five modules, transferred from the
    image-space masks in (a) by (b) intersecting the viewing ray of each contour vertex with the DSM,
    (c) a two-dimensional image-to-object affine transformation estimated for each candidate from DSM point correspondences and (d) per-module
    slope correction.}
    \label{fig:abl_projection}
\end{figure}

\subsection{Geometry-Guided Scoring and Cross-View Fusion}
\label{sec:postprocessing}

After slope correction, candidates from all images share one object-space coordinate system. Some still represent module fragments or merged groups, while others are repeated observations of the same module. We score their geometric plausibility before using NMS to select representative instances across views.

Rooftop PV arrays often use repeated module specifications and regular installation patterns. Representative dimensions can be obtained by measuring a module or consulting its product specification, providing a low-cost indication of the expected size of one instance. We use these dimensions as a soft constraint, not a requirement that every module have exactly the same size. The evaluated sites contain modest dimensional variation, which is accommodated through a tolerance in the score.

Let $L$ and $W$ be the representative module length and width, with nominal area $A_{\mathrm{ref}}=LW$. For a candidate with planimetric area $A$ and minimum rotated bounding-rectangle area $A_{\mathrm{rect}}$, the area score $s_A$ and rectangularity score $s_R$ are
\begin{align}
    s_A &= \exp\!\left(-\frac{|A-A_{\mathrm{ref}}|}{\delta_A A_{\mathrm{ref}}}\right), \label{eq:score_area}\\
    s_R &= \exp\!\left(-\frac{|A/A_{\mathrm{rect}}-1|}{\delta_R}\right), \label{eq:score_rect}
\end{align}
where the positive tolerances $\delta_A$ and $\delta_R$ control the decay of the two scores. The area score penalises fragments and merged groups that depart from the expected module size. The rectangularity score penalises irregular contours, including those with a plausible area. We combine them with the model confidence as
\begin{align}
    S_{\mathrm{geo}} &= \frac{w_A s_A+w_R s_R}{w_A+w_R}, \label{eq:score_geo}\\
    S\hphantom{_{\mathrm{geo}}} &= \frac{w_{\mathrm{conf}}s_{\mathrm{conf}}+w_{\mathrm{geo}}S_{\mathrm{geo}}}{w_{\mathrm{conf}}+w_{\mathrm{geo}}}, \label{eq:score_final}
\end{align}
where $S_{\mathrm{geo}}$ is the combined geometry score, $s_{\mathrm{conf}}\in[0,1]$ is the segmentation-model confidence, and $S$ is the final candidate score. The non-negative weights $w_A$ and $w_R$ balance the geometric terms, while $w_{\mathrm{conf}}$ and $w_{\mathrm{geo}}$ balance confidence and geometry; each pair has a positive sum. Candidates with $S<S_{\min}$ are discarded, where $S_{\min}$ is the acceptance threshold.

NMS processes the retained candidates in descending order of $S$ and suppresses overlapping lower-scoring candidates above a polygon-IoU threshold $\theta_{\mathrm{IoU}}$. Polygon overlap is used because axis-aligned bounding boxes can overlap substantially even for separate modules in a rotated array. An R-tree index limits overlap tests to spatially intersecting bounding boxes. The highest-scoring retained observation represents each module; this instance set supplies both the current map and the next round of prompts.

\subsection{Iterative Covering-View Feedback}
\label{sec:iterative-pipeline}

The iteration index $t$ counts feedback rounds, with $t=0$ denoting the initial text-only pass. Segmentation, slope correction, scoring and fusion yield the accepted instance set for that round. Although each retained polygon originates from one image, its role is not confined to that source view: it is projected into every image whose ground footprint intersects the instance. The image-space bounding box of each projected polygon is then used as a spatial prompt for re-segmentation.

The next round re-segments all windows using the prompting strategy in Section~\ref{sec:sam3}. The resulting candidates undergo projection, geometric verification and cross-view fusion again, updating the map and the evidence available for subsequent rounds. Routing each instance to all covering views allows a successful detection in one image to guide another image in which that module was previously missed or poorly delineated. The last accepted set is the output. We evaluate successive rounds separately in Section~\ref{sec:budget} to distinguish gains in module recovery from gains in boundary accuracy.

\section{Results and Discussion}
\label{sec:results}

The evaluation examines whether MV-PVMapper produces a complete module map with separate, geometrically consistent instances. We first compare MV-PVMapper with DOM-IterSeg and MV-PixelVote, then examine the three main components of the framework, the number of feedback iterations and the available camera directions.

\subsection{Configuration and Implementation Details}
\label{sec:configuration}

All methods use the same pretrained SAM~3 model without task-specific weight updates. The experiments process $1024\times1024$-pixel windows with 50\% overlap and use the text prompt \texttt{a PV panel}. The confidence threshold is 0.45, the minimum mask area is 500 pixels, and each window receives at most $N_b=5$ box prompts. Projected boxes with either side shorter than 50 pixels are discarded. The representative module dimensions are listed in Table~\ref{tab:sites}.

For geometric scoring, we set $\delta_A=0.25$, $\delta_R=0.1$, $(w_A,w_R)=(10,5)$ and $(w_{\mathrm{conf}},w_{\mathrm{geo}})=(0.2,0.8)$. The acceptance threshold is $S_{\min}=0.7$, and the polygon-IoU threshold for NMS is $\theta_{\mathrm{IoU}}=0.2$. Unless specified otherwise, the reported maps are obtained after two feedback rounds ($t=2$), following the initial text-only pass. The iteration experiment also reports the intermediate results at $t=0$ and $t=1$.

All experiments were run on one NVIDIA RTX 5880 Ada GPU (48\,GB) and 128\,GB of RAM, under Ubuntu 22.04.5 LTS with Python 3.12.12 and PyTorch 2.6.0 (CUDA 11.8). SAM~3 is used through its official implementation and released weights~\citep{carion2025sam}.

\subsection{Evaluation Metrics and Baseline Design}
\label{sec:metrics}

All methods are evaluated as georeferenced polygons in the same object-space coordinate system. We assess PV coverage first and individual module instances second. Metrics are computed separately for each site; overall results are unweighted averages of the three site-level values, reported as percentages to two decimal places.

\subsubsection{Area-Level Agreement}

Area-level evaluation follows the foreground-overlap principle used in semantic segmentation \citep{long2015fully}. Predicted polygons are merged into one coverage region, and reference polygons into another, without retaining instance identities. Area IoU is their intersection area divided by their union area. Area precision is the fraction of predicted coverage supported by the reference, area recall is the fraction of reference coverage recovered, and area $F_1$ is their harmonic mean. These metrics assess coverage but do not distinguish boundaries between neighbouring modules.

\subsubsection{Instance-Level Agreement}

We use recognition quality (RQ), segmentation quality (SQ) and panoptic quality (PQ) to separate instance recovery from boundary accuracy \citep{kirillov2019panoptic}. Predictions and references are matched one-to-one at an IoU threshold of 0.5. True positives (TP) are matched predictions, false positives (FP) are unmatched predictions, and false negatives (FN) are unmatched reference modules. Let $\mathcal{T}$ denote the set of matched pairs $(d,g)$, where $d$ is a predicted polygon and $g$ its reference polygon. Then
\begin{equation}
    \mathrm{RQ}=\frac{2\,\mathrm{TP}}{2\,\mathrm{TP}+\mathrm{FP}+\mathrm{FN}},\quad
    \mathrm{SQ}=\frac{1}{\mathrm{TP}}\sum_{(d,g)\in\mathcal{T}}\mathrm{IoU}(d,g),\quad
    \mathrm{PQ}=\mathrm{SQ}\times\mathrm{RQ}.
    \label{eq:pq}
\end{equation}
RQ is the harmonic mean of instance precision and recall. SQ measures the mean overlap of successfully matched modules and must therefore be read together with RQ. PQ combines both aspects.

We also report AJI, which aggregates overlap by instance rather than over the merged foreground \citep{kumar2017dataset}. For consistency across our instance metrics, we use the same thresholded one-to-one matching defined above:
\begin{equation}
    \mathrm{AJI}=\frac{\sum_{(d,g)\in\mathcal{T}}|d\cap g|}
    {\sum_{(d,g)\in\mathcal{T}}|d\cup g|+\sum_{u\in\mathcal{U}}|u|},
    \label{eq:aji}
\end{equation}
where $|\,\cdot\,|$ denotes polygon area and $\mathcal{U}$ contains all unmatched predicted and reference polygons. This matching convention is an adaptation of the original AJI definition, which assigns predictions by maximum overlap without the present 0.5 threshold. The denominator penalises unmatched regions as well as imperfect matched boundaries. Finally, $AP_{95}$ measures average precision under a stricter IoU threshold of 0.95, assessing the recovery of accurately delineated modules.

\subsubsection{Baseline Design}
\label{sec:baselines}

The comparison tests two alternative mapping strategies while retaining SAM~3 as the segmentation model. The baselines adapt established processing ideas to the present training-free setting; they are not reproductions of the cited systems or direct comparisons with those papers' published results.

\begin{itemize}
    \item \textbf{DOM-IterSeg: iterative segmentation on an orthophoto.} Extracting modules from a georeferenced orthomosaic is an established PV mapping route \citep{cardoso2024automated}. We implement it with iterative SAM~3 segmentation and the module-geometry prior on each site's DOM. This baseline tests how far prompted segmentation and geometric screening can proceed within one orthorectified representation, without cross-view feedback from the source images.
    \item \textbf{MV-PixelVote: multi-view pixel-level decision fusion.} Projection-based semantic mapping can combine predictions from multiple views at shared spatial locations \citep{lawin2017deep}. We adapt this decision-fusion idea by segmenting the same perspective images used by MV-PVMapper, transferring their masks to object space and combining them by pixel-level voting. Voting thresholds are tuned separately for each site. This baseline uses multi-view evidence but aggregates foreground support at individual pixels rather than selecting module instances using their geometry.
\end{itemize}

MV-PVMapper uses the original perspective images, camera geometry and DSM; it does not use the DOM as a segmentation input. Thus, DOM-IterSeg assesses the orthophoto route, whereas MV-PixelVote assesses pixel-level aggregation of the original multi-view predictions. All three methods run two feedback rounds after the initial text-only pass and are compared at $t=2$. In each feedback round, DOM-IterSeg uses the bounding boxes of the previous round's polygons, converted to DOM pixel coordinates, as box prompts. MV-PixelVote counts the projected polygons covering each cell of a 0.02\,m object-space grid. Cells with at least 20, 27 and 30 votes at Sites~1, 2 and~3, respectively, are retained, and their connected components are converted to polygons. Polygons smaller than 0.6\,m$^2$ are removed.

\subsection{Main Results and Cross-Site Comparison}
\label{sec:main_results}

Table~\ref{tab:main_comparison} compares the three mapping strategies. MV-PVMapper achieves an AJI of 97.55\% and an $AP_{95}$ of 95.67\%, combining near-complete instance recovery with accurate boundaries. The baselines reveal two different limitations. DOM-IterSeg retains high area precision but recovers only 60.77\% of the reference area. MV-PixelVote recovers more coverage, yet its AJI falls to 41.24\% because foreground agreement does not ensure that neighbouring modules remain separate.
\par

\begin{table}[!htp]
\centering
\caption{Module-mapping accuracy of MV-PVMapper and the two baselines at $t=2$.
Values (\%) are macro-averages over the three sites. Higher values are better; bold marks the highest
value in each column.}
\label{tab:main_comparison}
\resizebox{\textwidth}{!}{
\begin{tabular}{lccccccccc}
\toprule
 & \multicolumn{4}{c}{\textbf{Area-level}} & \multicolumn{5}{c}{\textbf{Instance-level}} \\
\cmidrule(lr){2-5}\cmidrule(lr){6-10}
\textbf{Method} & IoU$_\mathrm{area}$ & $F_{1}$ & P$_\mathrm{area}$ & R$_\mathrm{area}$ & RQ & SQ & PQ & AJI & $AP_{95}$ \\
\midrule
DOM-IterSeg & 60.73 & 72.58 & \textbf{99.80} & 60.77 & 73.93 & 96.97 & 71.66 & 60.73 & 60.51 \\
MV-PixelVote            & 86.11 & 92.19 & 96.79 & 89.11 & 49.40 & 96.03 & 47.64 & 41.24 & 27.52 \\
\rowcolor{gray!10} \textbf{MV-PVMapper} & \textbf{97.55} & \textbf{98.76} & 99.00 & \textbf{98.52} & \textbf{99.99} & \textbf{97.56} & \textbf{97.56} & \textbf{97.55} & \textbf{95.67} \\
\bottomrule
\end{tabular}}
\end{table}

\FloatBarrier

\begin{figure}[H]
    \centering
    \includegraphics[width=\textwidth]{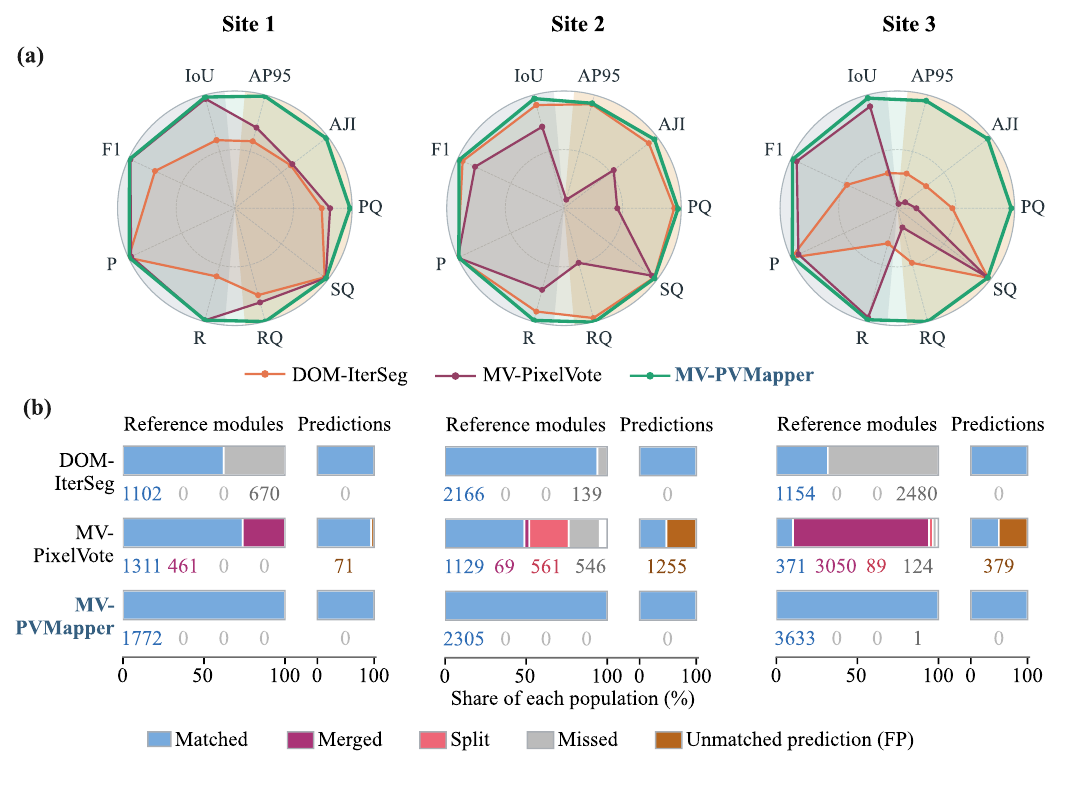}
    \caption{Per-site comparison of the three methods at $t=2$. (a) Metric profiles; IoU, F1, P and R
    are the area-level metrics of Table~\ref{tab:main_comparison}, and the dashed circle marks 50\%.
    (b) Reference-module outcomes and prediction-side false positives. Reference and prediction bars
    are normalised by their respective populations. Numbers below the reference bars give matched,
    merged, split and missed counts; numbers below the prediction bars give unmatched predictions (FP).}
    \label{fig:per_site}
\end{figure}

Figure~\ref{fig:per_site}a shows that DOM-IterSeg retains high area precision but has much lower area recall at Sites~1 and~3, whereas MV-PixelVote has comparatively high area IoU and $F_1$ but markedly lower RQ, AJI and $AP_{95}$ at Sites~2 and~3. The reference-module bars in Figure~\ref{fig:per_site}b explain these contrasting profiles: DOM-IterSeg misses 670 and 2\,480 modules at Sites~1 and~3, respectively, while MV-PixelVote merges or splits many reference modules, including 461 merged modules at Site~1 and 561 split modules at Site~2. The separate prediction bars also reveal surplus MV-PixelVote predictions, particularly at Site~2. MV-PVMapper maintains high scores across all three sites, with almost entirely matched reference-module bars and no unmatched predictions.
\par

\FloatBarrier

\begin{evidencewithtext}
Across all three sites, MV-PVMapper matches 7\,710 of the 7\,711 reference modules, with no false positives and no merged or split modules. The site-wide maps in Figure~\ref{fig:main_global} show the different spatial patterns of baseline errors: DOM-IterSeg leaves scattered gaps at Site~1 and misses large portions of the three roof-wide arrays at Site~3, while MV-PixelVote covers more of these arrays but merges neighbouring modules, including 3\,050 reference modules at Site~3.
\par

\begin{figure}[H]
    \centering
    \includegraphics[width=0.90\textwidth]{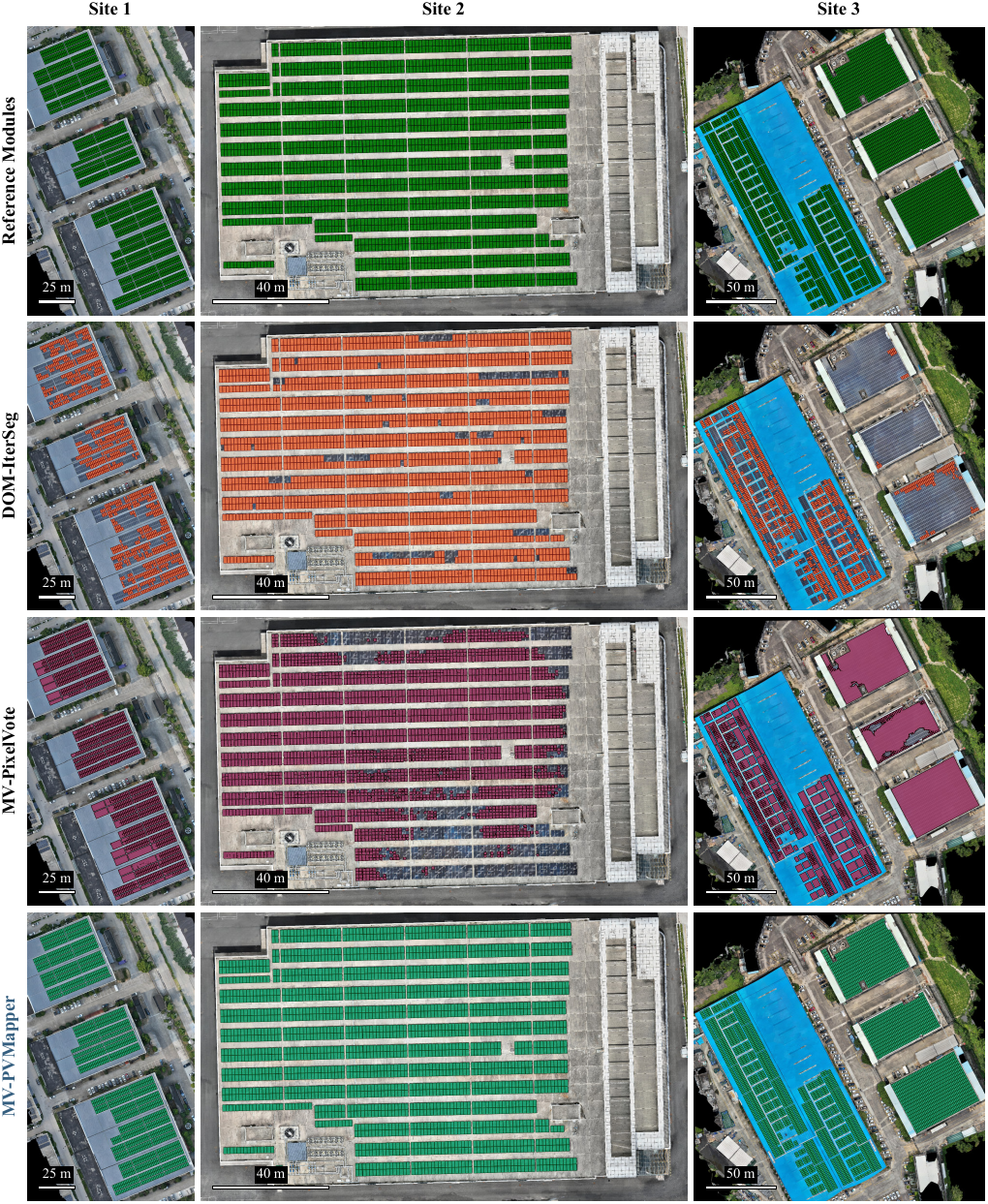}
    \caption{Site-wide module maps at $t=2$, overlaid on the DOM. Columns show Sites~1--3 and rows
    show the reference modules, DOM-IterSeg, MV-PixelVote and MV-PVMapper. }
    \label{fig:main_global}
\end{figure}

\end{evidencewithtext}
\FloatBarrier

\begin{evidencewithtext}
The paired crops in Figure~\ref{fig:qualitative_results} make the instance errors visible: MV-PixelVote fills narrow inter-module gaps at Sites~1 and~3 and produces fragmented predictions at Site~2, where DOM-IterSeg retains mostly separate modules. MV-PVMapper preserves both the site-wide array pattern and the individual module contours, showing why foreground coverage alone is insufficient to assess module mapping.
\par

\begin{figure}[H]
    \centering
    \includegraphics[width=\textwidth]{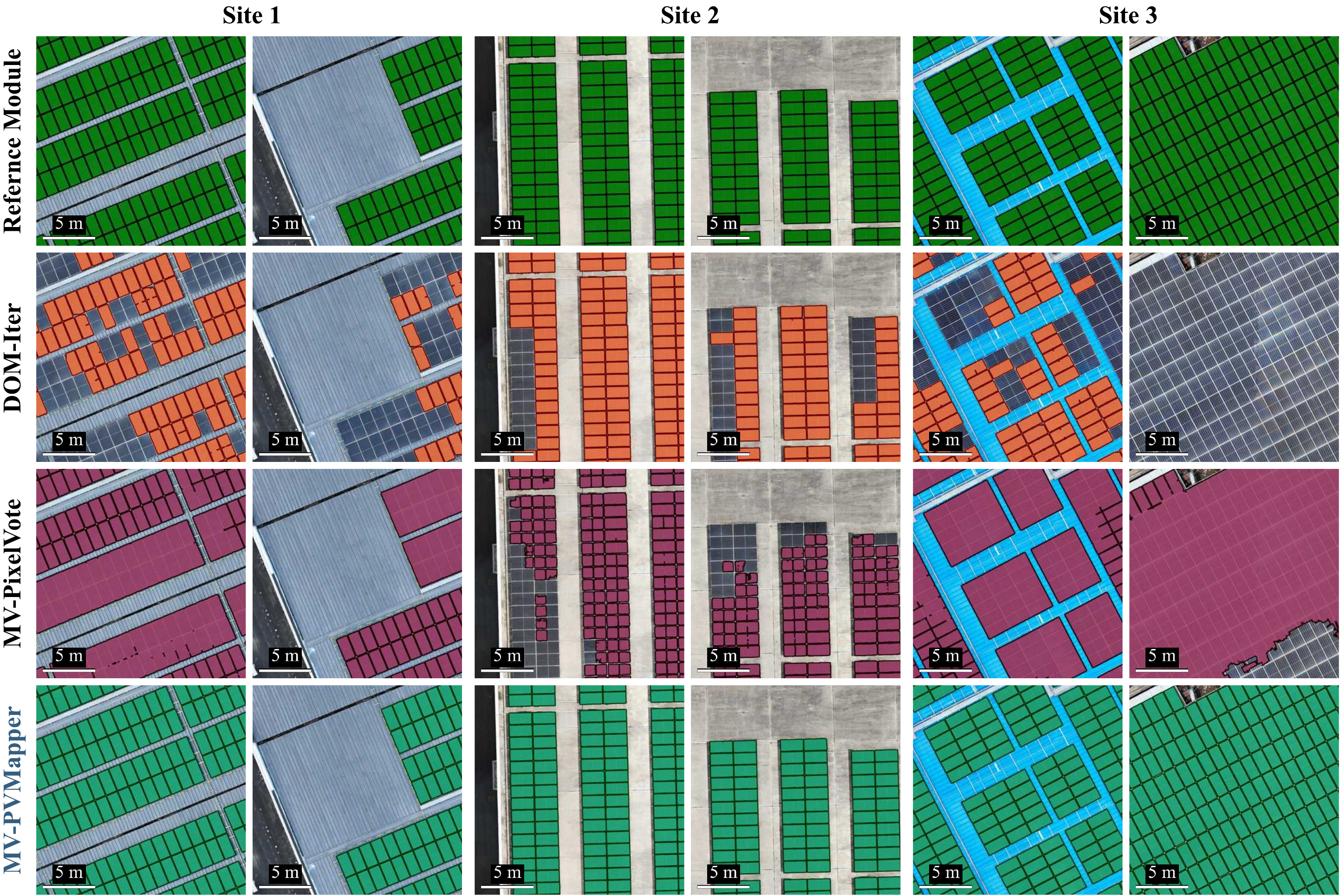}
    \caption{Detail views of the module maps in Figure~\ref{fig:main_global}. Each site is shown by
    two 5\,m-scale crops; rows show the reference modules, DOM-IterSeg, MV-PixelVote and MV-PVMapper.}
    \label{fig:qualitative_results}
\end{figure}

\end{evidencewithtext}
\FloatBarrier

\subsection{Ablation Studies}
\label{sec:ablation}

The ablations isolate contour projection, the module-geometry prior and feedback routing. Each changes one component of the full framework while retaining the remaining configuration. In each ablation, the configuration labelled ours corresponds to the MV-PVMapper result in Table~\ref{tab:main_comparison}. Results are evaluated after two feedback rounds, except for the no-feedback control, and the tables report macro-averages over the three sites.
\par

\subsubsection{Contour Projection}
\label{sec:ablation-projection}

This experiment tests whether a module-specific surface model improves contour transfer and subsequent instance selection. We compare:
\par
\begin{itemize}
    \item \textbf{Collinearity-based projection:} intersect each contour vertex's viewing ray directly with the DSM.
    \item \textbf{Affine-based projection:} fit a two-dimensional affine transformation from DSM-derived image--object correspondences within each candidate's image bounding box. This adapts the affine-mapping idea of MVMapper \citep{chen2024mvmapper} to individual candidates.
    \item \textbf{Slope correction projection (ours):} fit a local module plane from visible DSM points that project into the mask and intersect the contour rays with that plane.
\end{itemize}

\begin{evidencewithtext}
Table~\ref{tab:abl_projection} shows that slope correction projection reaches 97.55\% area IoU and 95.67\% $AP_{95}$, compared with 97.24\% and 93.46\% for affine-based projection and 89.94\% and 46.56\% for collinearity-based projection.
\par

\begin{table}[H]
\centering
\caption{Ablation of contour projection. All configurations use the full geometry prior and
feedback to all covering views and are evaluated at $t=2$. Values (\%) are macro-averages over the
three sites; bold marks the highest value in each column.}
\label{tab:abl_projection}
\ifdefined\ablfirstcolw\else\newlength{\ablfirstcolw}\fi
\settowidth{\ablfirstcolw}{\textbf{Object-space covering views (ours)}}
\resizebox{\textwidth}{!}{
\begin{tabular}{lccccccccc}
\toprule
 & \multicolumn{4}{c}{\textbf{Area-level}} & \multicolumn{5}{c}{\textbf{Instance-level}} \\
\cmidrule(lr){2-5}\cmidrule(lr){6-10}
\makebox[\ablfirstcolw][l]{\textbf{Projection}} & IoU$_\mathrm{area}$ & $F_1$ & P$_\mathrm{area}$ & R$_\mathrm{area}$ & RQ & SQ & PQ & AJI & $AP_{95}$ \\
\midrule
Collinearity-based projection & 89.94 & 94.52 & 96.78 & 92.56 & 96.85 & 94.45 & 91.54 & 87.70 & 46.56 \\
Affine-based projection       & 97.24 & 98.60 & 98.82 & 98.38 & 99.98 & 97.28 & 97.26 & 97.24 & 93.46 \\
\rowcolor{gray!10} \textbf{Slope correction projection (ours)} & \textbf{97.55} & \textbf{98.76} & \textbf{99.00} & \textbf{98.52} & \textbf{99.99} & \textbf{97.56} & \textbf{97.56} & \textbf{97.55} & \textbf{95.67} \\
\bottomrule
\end{tabular}}
\end{table}

\end{evidencewithtext}
\FloatBarrier

\begin{evidencewithtext}
Figure~\ref{fig:abl_proj_analysis}a--c shows that collinearity-based projection mainly reduces boundary accuracy (lower SQ and $AP_{95}$) at Sites~1 and~3. Site~2 additionally exhibits a pronounced detection failure: 15.44\% of reference modules are missed and 2.41\% of predictions are false positives, with RQ falling to 90.56\% and $AP_{95}$ to 19.82\%. The geometric reasons for this site-specific failure are further examined through the module-level 3D geometry and elevation profiles in Figure~\ref{fig:3d_tile}. Slope correction recovers all modules at Site~2 without false positives and raises $AP_{95}$ to 93.20\%.
\par

\begin{figure}[H]
    \centering
    \includegraphics[width=\textwidth]{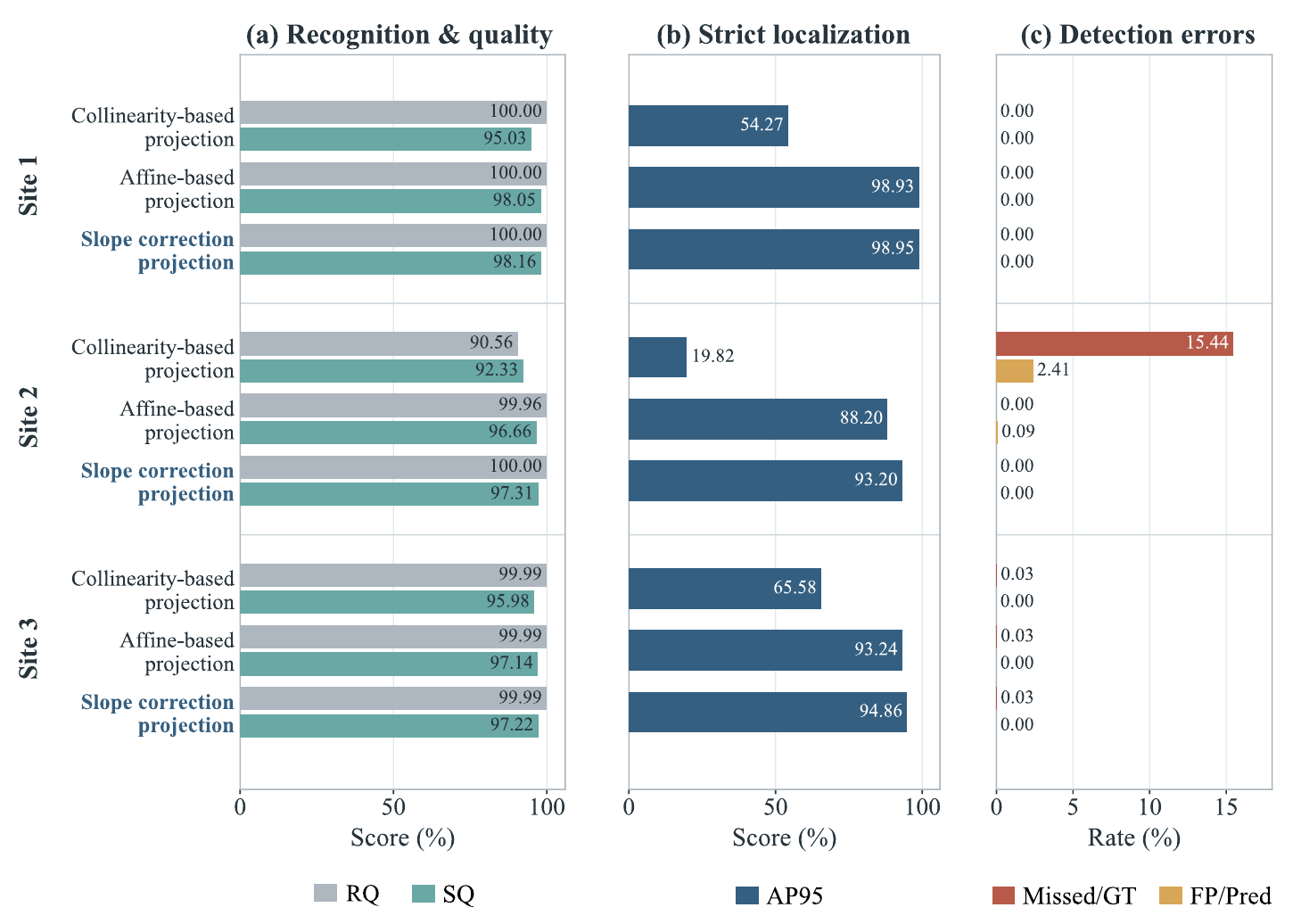}
    \caption{Per-site results of the contour-projection ablation at $t=2$. (a) RQ
    and SQ; (b) $AP_{95}$; (c) missed reference modules (Missed/GT) and false positives among the
    predictions (FP/Pred).}
    \label{fig:abl_proj_analysis}
\end{figure}

\end{evidencewithtext}
\FloatBarrier

\begin{evidencewithtext}
Figure~\ref{fig:abl_proj_detail} distinguishes boundary discrepancies from whole-module omissions. At Sites~1 and~3, FP and FN pixels are concentrated along module perimeters, while most module interiors remain correctly covered. At Site~2, the collinearity-based result contains entire module interiors labelled as FN, whereas affine-based and slope correction projection recover these regions. The residual differences between the latter two methods occur mainly in narrow boundary bands, illustrating why similar instance-recovery rates can coexist with differences in $AP_{95}$.
\par

\begin{figure}[H]
    \centering
    \includegraphics[width=\textwidth]{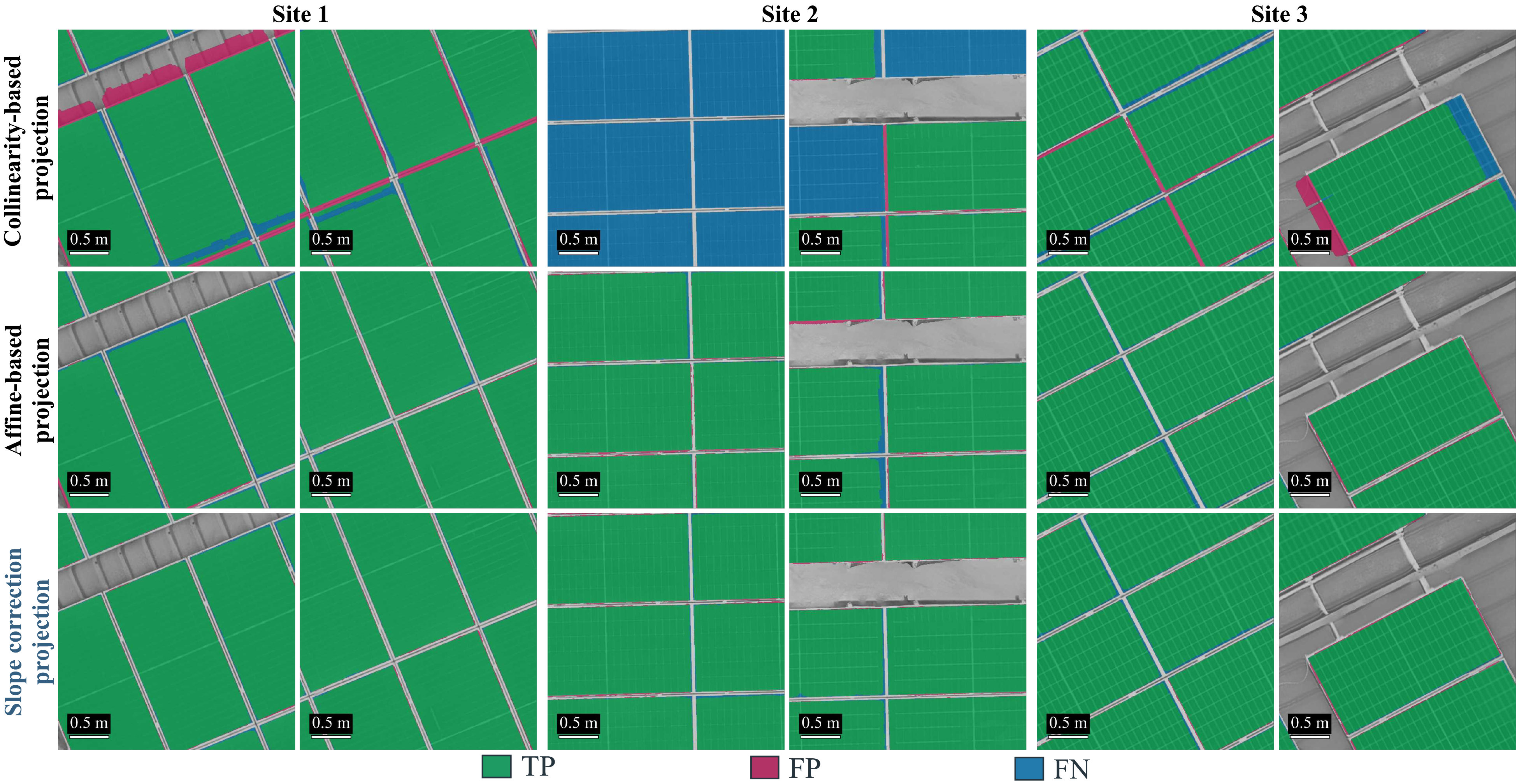}
    \caption{Object-space contours obtained with the three projection methods (rows) in two windows
    per site (columns); TP, FP and FN are shown at the pixel level.}
    \label{fig:abl_proj_detail}
\end{figure}

\end{evidencewithtext}
\FloatBarrier

\begin{evidencewithtext}
The module-level 3D profiles in Figure~\ref{fig:3d_tile} complement the quantitative results and object-space contours by showing the installation geometry underlying the differences between sites. At Sites~1 and~3, modules lie close to the roof surface. At Site~2, each rack-mounted row has an upper edge approximately 1.5\,m above the roof, producing an abrupt height transition. Collinearity-based projection can then distort an otherwise valid candidate enough to reduce its geometry score. Consistent with this mechanism, it misses 356 modules at Site~2, whereas slope correction projection matches all 2\,305.
\par

\begin{figure}[H]
    \centering
    \includegraphics[width=\textwidth]{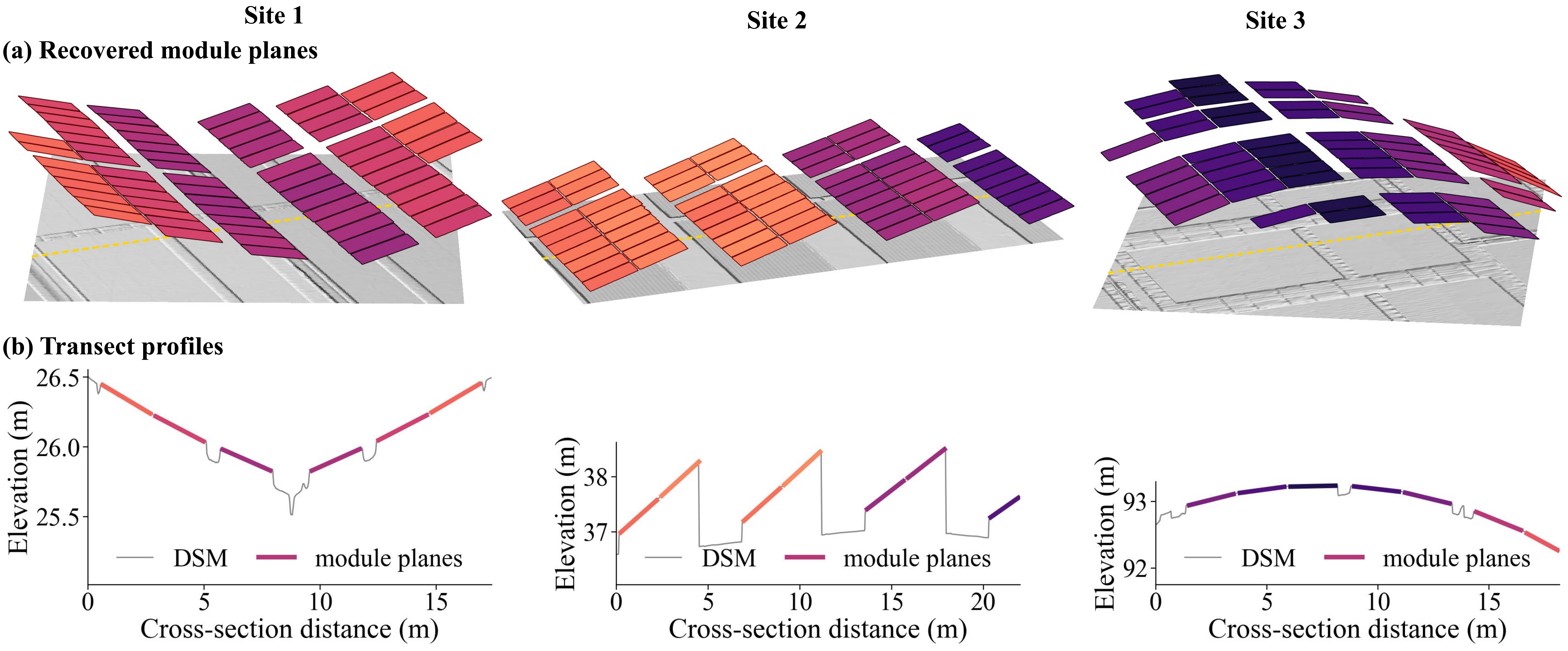}
    \caption{Module-level 3D geometry at Sites~1--3 (columns). (a) Recovered module planes above the
    DSM, coloured by tilt on a separate scale for each site; the yellow dashed line marks the
    transect. (b) Elevation of the DSM and the module planes along the transect.}
    \label{fig:3d_tile}
\end{figure}

\end{evidencewithtext}

\FloatBarrier
\subsubsection{Module-Geometry Prior}
\label{sec:ablation-prior}

This experiment separates the roles of size and shape in deciding whether a candidate represents one module. The four configurations are:
\par
\begin{itemize}
    \item \textbf{No geometry prior:} use segmentation confidence without the area or rectangularity term.
    \item \textbf{Rectangularity only (+\,rect):} retain the rectangularity term but remove the area term.
    \item \textbf{Area only (+\,area):} retain the area term but remove the rectangularity term.
    \item \textbf{Full prior (+\,rect +\,area; ours):} combine both terms as in MV-PVMapper.
\end{itemize}

\begin{evidencewithtext}
Without the geometry prior, area recall remains 98.09\%, but RQ falls to 5.16\% and AJI to 19.63\% (Table~\ref{tab:abl_prior}). The area term supplies the missing instance-scale constraint: on its own, it restores RQ to 99.95\%. Combining it with rectangularity further improves $AP_{95}$ from 90.81\% to 95.67\%.
\par

\begin{table}[H]
\centering
\caption{Ablation of the module-geometry prior. +\,rect and +\,area each use only the corresponding
term. All configurations use slope correction and feedback to all covering views and are evaluated
at $t=2$. Values (\%) are macro-averages over the three sites; bold marks the highest value in each
column.}
\label{tab:abl_prior}
\ifdefined\ablfirstcolw\else\newlength{\ablfirstcolw}\fi
\settowidth{\ablfirstcolw}{\textbf{Object-space covering views (ours)}}
\resizebox{\textwidth}{!}{
\begin{tabular}{lccccccccc}
\toprule
 & \multicolumn{4}{c}{\textbf{Area-level}} & \multicolumn{5}{c}{\textbf{Instance-level}} \\
\cmidrule(lr){2-5}\cmidrule(lr){6-10}
\makebox[\ablfirstcolw][l]{\textbf{Geometry prior}} & IoU$_\mathrm{area}$ & $F_1$ & P$_\mathrm{area}$ & R$_\mathrm{area}$ & RQ & SQ & PQ & AJI & $AP_{95}$ \\
\midrule
No geometry prior             & 72.99 & 84.38 & 74.05 & 98.09 & 5.16 & 91.91 & 4.84 & 19.63 & 3.80 \\
+\,rect                       & 95.34 & 97.60 & 98.18 & 97.04 & 84.03 & 97.09 & 81.64 & 83.69 & 87.54 \\
+\,area                       & 97.20 & 98.58 & 98.63 & \textbf{98.53} & 99.95 & 97.29 & 97.24 & 97.19 & 90.81 \\
\rowcolor{gray!10} \textbf{+\,rect +\,area (ours)} & \textbf{97.55} & \textbf{98.76} & \textbf{99.00} & 98.52 & \textbf{99.99} & \textbf{97.56} & \textbf{97.56} & \textbf{97.55} & \textbf{95.67} \\
\bottomrule
\end{tabular}}
\end{table}

\end{evidencewithtext}
\FloatBarrier

\begin{evidencewithtext}
Without the geometry prior, Figure~\ref{fig:abl_prior_analysis}a shows prediction counts 5.82--7.21 times the reference counts despite area $F_1$ remaining above 83\% at all three sites. Panels (b)--(c) reveal low RQ and AJI alongside extensive merging and splitting, whereas the area term brings prediction counts close to the reference counts and eliminates the reported partition errors. Adding rectangularity to the area term further increases $AP_{95}$ at every site, indicating an additional benefit for boundary accuracy beyond recovering the correct number of instances.
\par

\begin{figure}[H]
    \centering
    \includegraphics[width=\textwidth]{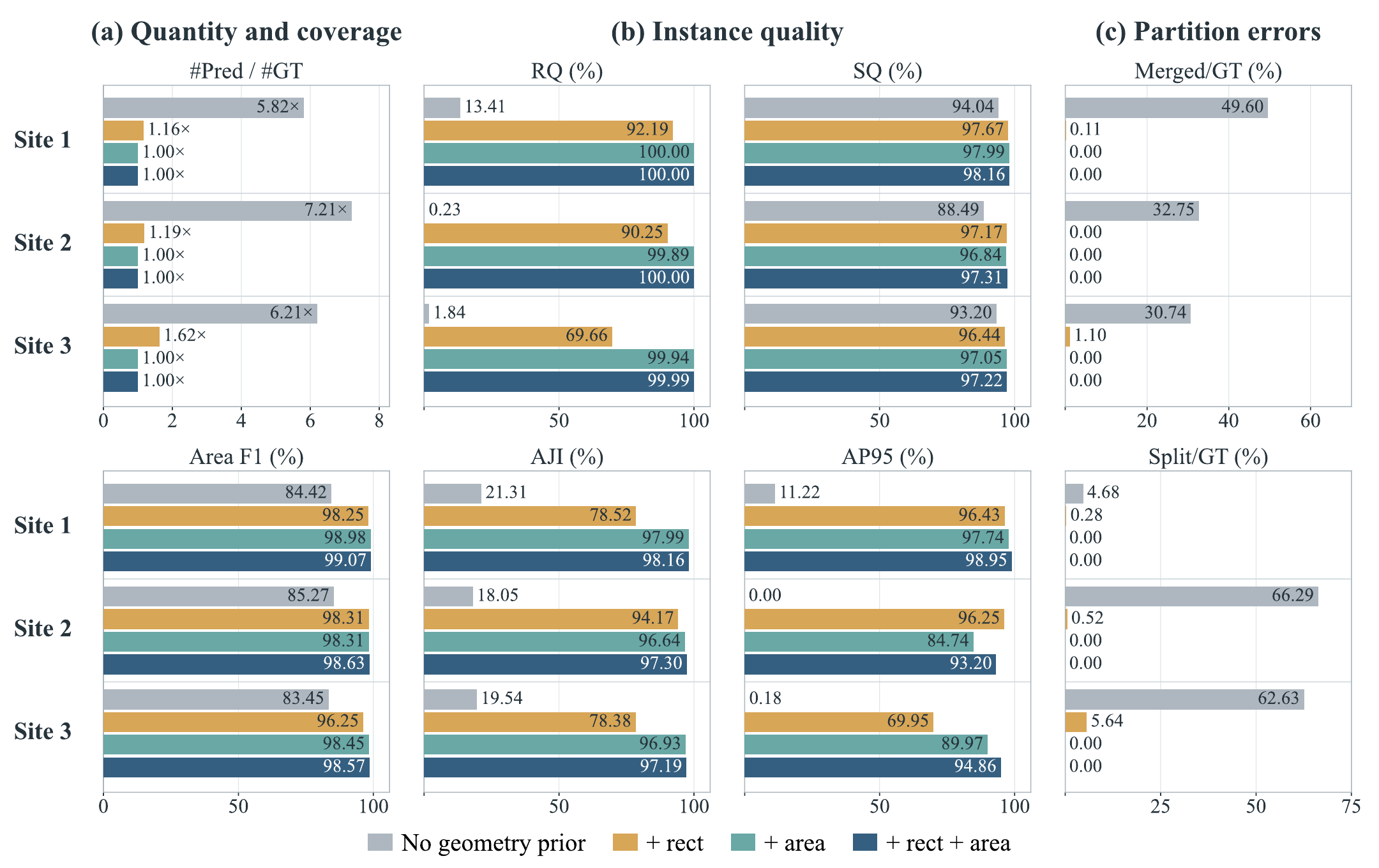}
    \caption{Per-site results of the module-geometry prior ablation at $t=2$. (a) Quantity and
    coverage: number of predictions relative to the number of reference modules (\#Pred/\#GT) and area $F_1$. (b) Instance quality: RQ and SQ (top),
    AJI and $AP_{95}$ (bottom). (c) Partition errors: merged and split reference modules as
    percentages of the reference count.}
    \label{fig:abl_prior_analysis}
\end{figure}

\end{evidencewithtext}
\FloatBarrier

\begin{evidencewithtext}
Rectangularity alone removes irregular candidates but cannot distinguish one module from a rectangular fragment or a rectangle spanning several modules. The examples in Figure~\ref{fig:abl_prior_detail} show these scale errors remaining on the arrays despite high area precision.
\par

\begin{figure}[H]
    \centering
    \includegraphics[width=\textwidth]{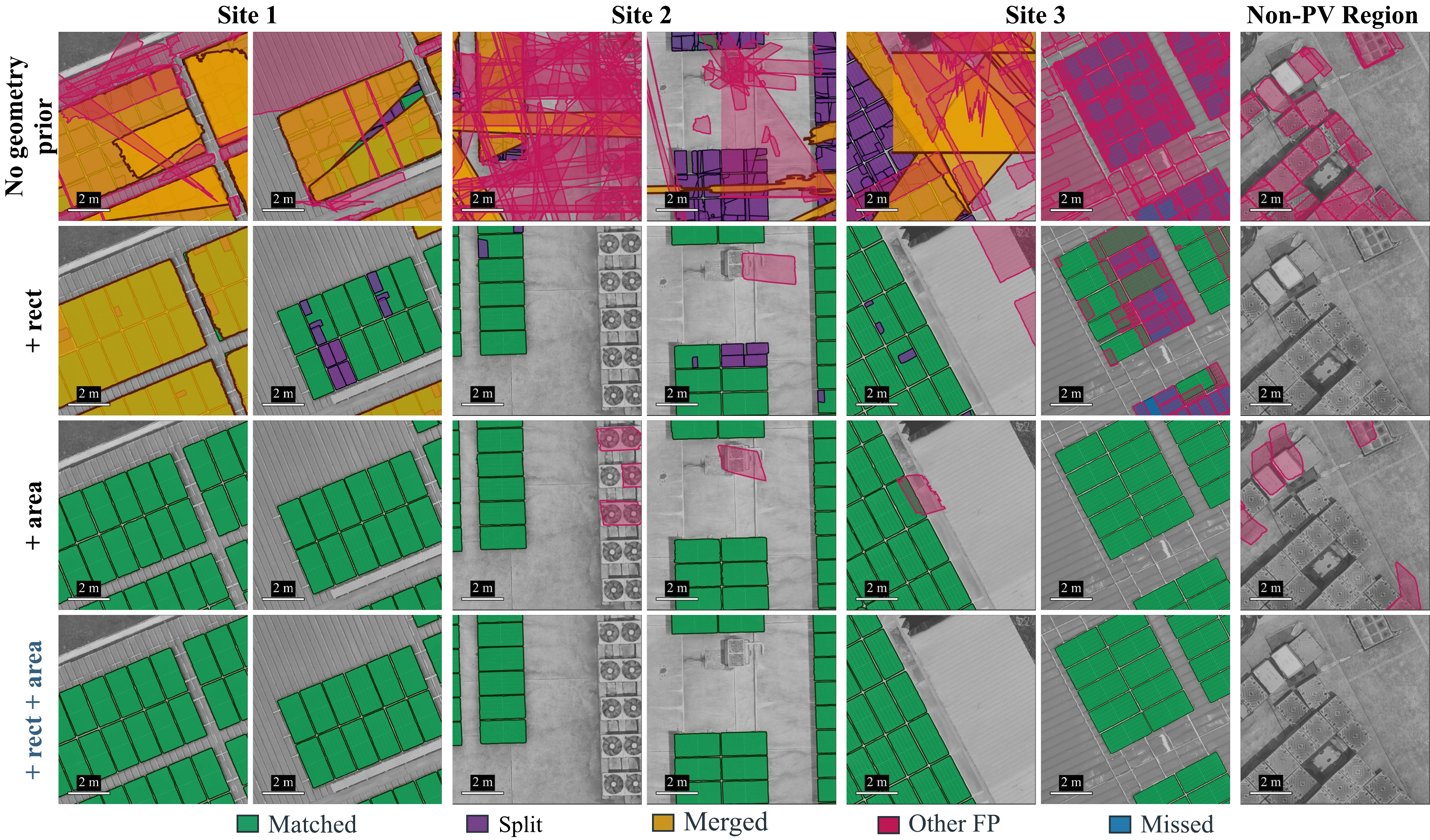}
    \caption{Module maps of the four geometry-prior configurations (rows) in two windows per site and
    one window over a non-PV area (columns). Predictions are coloured by outcome as shown in the
    legend.}
    \label{fig:abl_prior_detail}
\end{figure}

\end{evidencewithtext}
\FloatBarrier

The two terms therefore have distinct roles. Area is central to recovering the correct module inventory, while rectangularity improves the selection of well-delineated candidates at the expected scale. Neither strong coverage metrics nor the quality of matched pairs alone adequately describes the failures observed when these constraints are removed.
\par

\FloatBarrier
\subsubsection{Feedback Routing}
\label{sec:ablation-feedback}

This experiment asks whether feedback benefits from cross-view evidence, rather than simply from repeated model calls. We compare four ways of generating and routing prompts:
\par
\begin{itemize}
    \item \textbf{No feedback:} retain the initial text-only result at $t=0$.
    \item \textbf{Image-space self re-prompting:} derive each image's boxes from its own predictions in the previous round, without using the fused object-space result.
    \item \textbf{Source-view routing:} return each accepted object-space instance only to the image that supplied its representative polygon.
    \item \textbf{Covering-view routing (ours):} return each accepted object-space instance to all images that cover its location.
\end{itemize}

\begin{evidencewithtext}
Table~\ref{tab:abl_feedback} shows that the restricted routes improve instance recovery but do not match the boundary accuracy of covering-view routing. Image-space self re-prompting reaches a higher RQ than source-view routing, yet its SQ remains slightly below the no-feedback result. Their $AP_{95}$ values are 79.88\% and 86.77\%, compared with 95.67\% for MV-PVMapper. Thus, repeated prompting or returning a fused result only to its source image does not provide the full benefit of cross-view feedback.
\par

\begin{table}[H]
\centering
\caption{Ablation of feedback routing. No feedback is evaluated at $t=0$ and the other
configurations at $t=2$. All configurations use slope correction and the full
geometry prior. Values (\%) are macro-averages over the three sites; bold marks the highest value
in each column.}
\label{tab:abl_feedback}
\ifdefined\ablfirstcolw\else\newlength{\ablfirstcolw}\fi
\settowidth{\ablfirstcolw}{\textbf{Object-space covering views (ours)}}
\resizebox{\textwidth}{!}{
\begin{tabular}{lccccccccc}
\toprule
 & \multicolumn{4}{c}{\textbf{Area-level}} & \multicolumn{5}{c}{\textbf{Instance-level}} \\
\cmidrule(lr){2-5}\cmidrule(lr){6-10}
\makebox[\ablfirstcolw][l]{\textbf{Feedback routing}} & IoU$_\mathrm{area}$ & $F_1$ & P$_\mathrm{area}$ & R$_\mathrm{area}$ & RQ & SQ & PQ & AJI & $AP_{95}$ \\
\midrule
No feedback ($t=0$)           & 83.19 & 89.52 & 98.83 & 83.99 & 91.05 & 96.67 & 88.04 & 83.18 & 71.96 \\
Image-space self re-prompting & 95.36 & 97.62 & 98.74 & 96.53 & 99.35 & 96.53 & 95.90 & 95.32 & 79.88 \\
Source-view routing           & 94.80 & 97.30 & 98.84 & 95.86 & 98.77 & 97.05 & 95.86 & 94.80 & 86.77 \\
\rowcolor{gray!10} \textbf{Covering-view routing (ours)} & \textbf{97.55} & \textbf{98.76} & \textbf{99.00} & \textbf{98.52} & \textbf{99.99} & \textbf{97.56} & \textbf{97.56} & \textbf{97.55} & \textbf{95.67} \\
\bottomrule
\end{tabular}}
\end{table}

\end{evidencewithtext}
\FloatBarrier

\begin{evidencewithtext}
Figure~\ref{fig:abl_fb_analysis} distinguishes the two effects by site. Sites~1 and~2 already recover all modules in the initial pass, so feedback mainly improves their boundaries. At Site~3, self re-prompting and source-view routing still miss 117 and 258 modules, respectively, whereas covering-view routing misses only one.
\par

\begin{figure}[H]
    \centering
    \includegraphics[width=\textwidth]{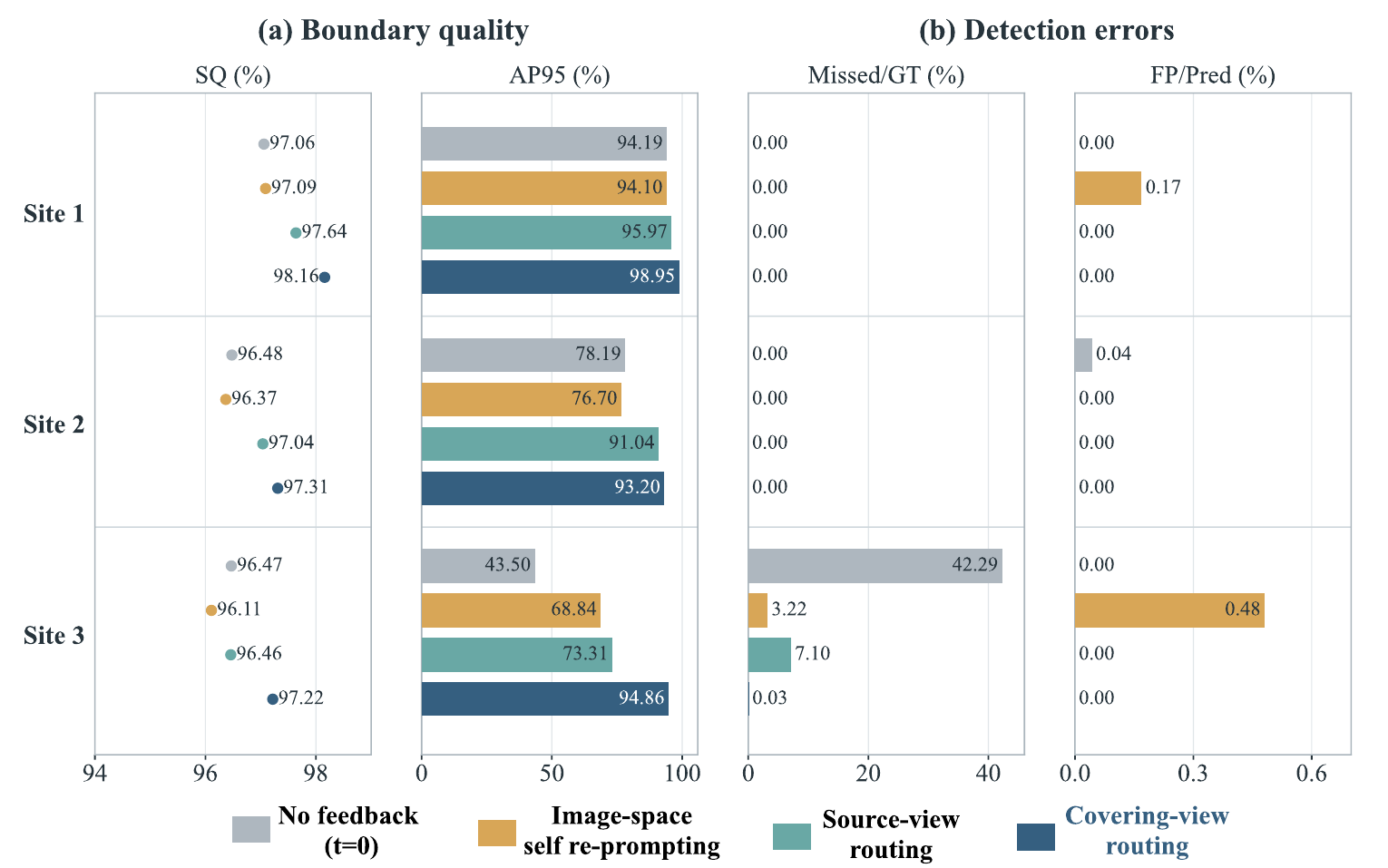}
    \caption{Per-site results of the feedback-routing ablation. No feedback is evaluated at $t=0$ and
    the other routes at $t=2$. (a) SQ and $AP_{95}$; (b) missed reference modules
    (Missed/GT) and false positives among the predictions (FP/Pred).}
    \label{fig:abl_fb_analysis}
\end{figure}

\end{evidencewithtext}
\FloatBarrier

\begin{evidencewithtext}
Figure~\ref{fig:abl_fb_detail} shows the corresponding residual gaps and boundary slivers. Sending verified instances to images that did not initially detect them is therefore important for both recovering dense arrays and refining their contours.
\par

\begin{figure}[H]
    \centering
    \includegraphics[width=\textwidth]{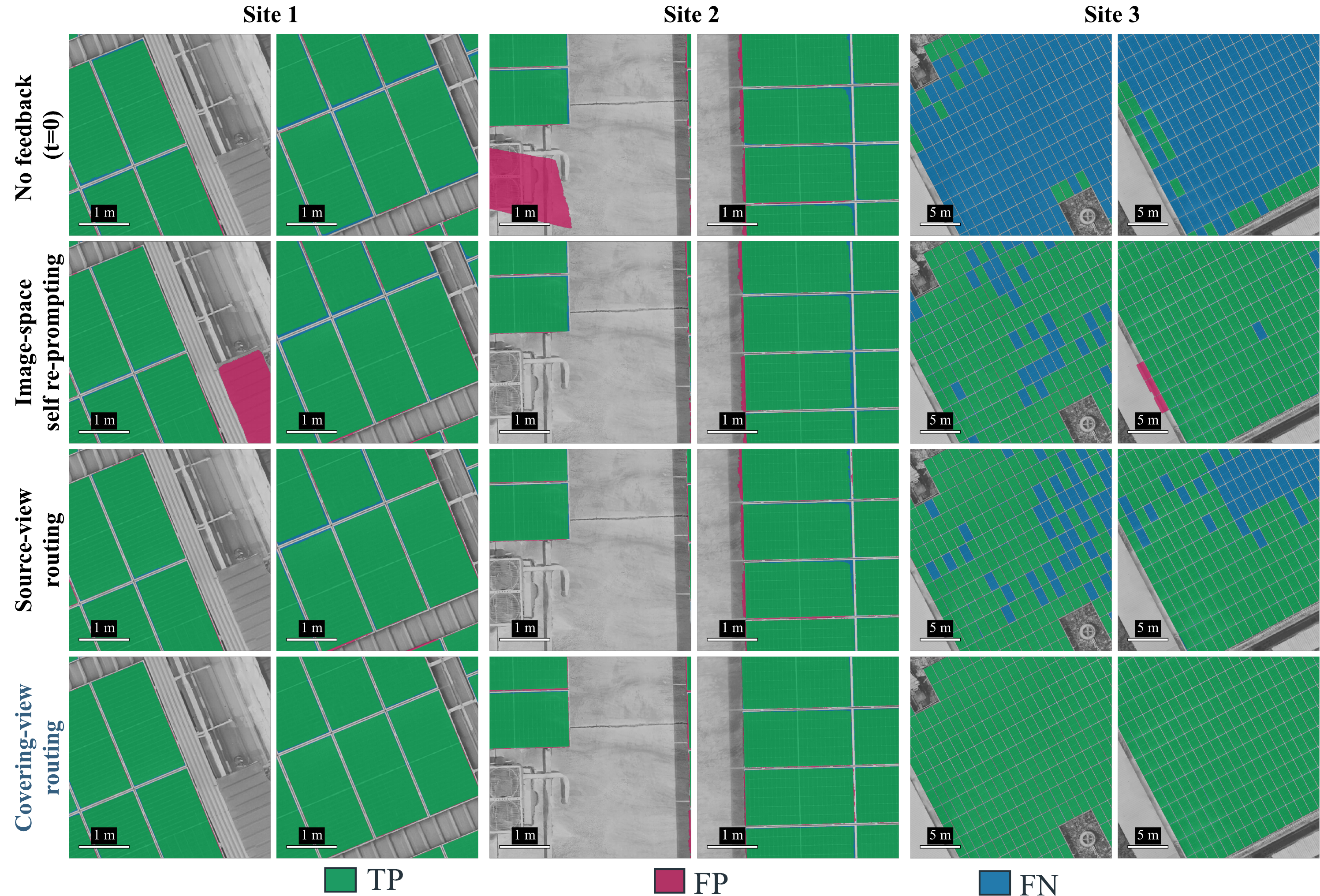}
    \caption{Comparison of the four feedback routes (rows) in two windows per site (columns); TP, FP
    and FN are shown at the pixel level.}
    \label{fig:abl_fb_detail}
\end{figure}

\end{evidencewithtext}

\FloatBarrier
\subsection{Iteration Count and Viewpoint Configuration}
\label{sec:budget}

This section examines how the number of feedback iterations and the number of camera directions affect MV-PVMapper. The iteration experiment uses the complete five-direction image set and evaluates the results at $t=0$, $1$ and $2$; the $t=2$ result is the main-experiment MV-PVMapper configuration in Table~\ref{tab:main_comparison}. The viewpoint experiment uses nested image subsets that start from the nadir images and add the four oblique directions one at a time in a fixed order. All subsets use camera poses and the DSM reconstructed from the complete five-direction survey; only the images used for segmentation vary.
\par

\begin{evidencewithtext}
Feedback both recovers missed modules and improves boundary accuracy (Table~\ref{tab:iteration_budget}). From $t=0$ to $t=2$, the number of missed modules summed over the three sites falls from 1\,537 to 125 and then to one, and the macro-averaged $AP_{95}$ rises by 23.71 percentage points, while area precision remains above 98.8\%. With a single missed module but an AJI of 97.55\% at $t=2$, the remaining error lies mainly at the module boundaries. The box prompts of each iteration mark every accepted module in all covering images, so a module that was missed or poorly delineated in one view but accepted from others is segmented again in that view; detection and boundary accuracy therefore improve together.
\par

\begin{table}[H]
\centering
\caption{Module-mapping accuracy at $t=0$, $1$ and $2$ using all five viewing directions. The $t=2$ row corresponds to MV-PVMapper in Table~\ref{tab:main_comparison}. Values (\%) are macro-averages over the three
sites; bold marks the highest value in each column.}
\label{tab:iteration_budget}
\resizebox{\textwidth}{!}{
\begin{tabular}{lccccccccc}
\toprule
 & \multicolumn{4}{c}{\textbf{Area-level}} & \multicolumn{5}{c}{\textbf{Instance-level}} \\
\cmidrule(lr){2-5}\cmidrule(lr){6-10}
\textbf{Iteration} & IoU$_\mathrm{area}$ & $F_1$ & P$_\mathrm{area}$ & R$_\mathrm{area}$ & RQ & SQ & PQ & AJI & $AP_{95}$ \\
\midrule
$t=0$ & 83.19 & 89.52 & 98.83 & 83.99 & 91.05 & 96.67 & 88.04 & 83.18 & 71.96 \\
$t=1$ & 96.33 & 98.12 & \textbf{99.01} & 97.26 & 99.42 & 97.43 & 96.86 & 96.33 & 92.44 \\
\rowcolor{gray!10} $t=2$ & \textbf{97.55} & \textbf{98.76} & 99.00 & \textbf{98.52} & \textbf{99.99} & \textbf{97.56} & \textbf{97.56} & \textbf{97.55} & \textbf{95.67} \\
\bottomrule
\end{tabular}}
\end{table}

\end{evidencewithtext}
\FloatBarrier

\begin{evidencewithtext}
Figure~\ref{fig:iter_by_site} shows that the two effects occur at different sites. Sites~1 and~2 have no missed modules at any iteration, so feedback acts there mainly on the boundaries: $AP_{95}$ rises by 4.76 and 15.01 percentage points, almost entirely in the first iteration. The second iteration does not change the recovered module counts at these two sites.
\par

\begin{figure}[H]
    \centering
    \includegraphics[width=\textwidth]{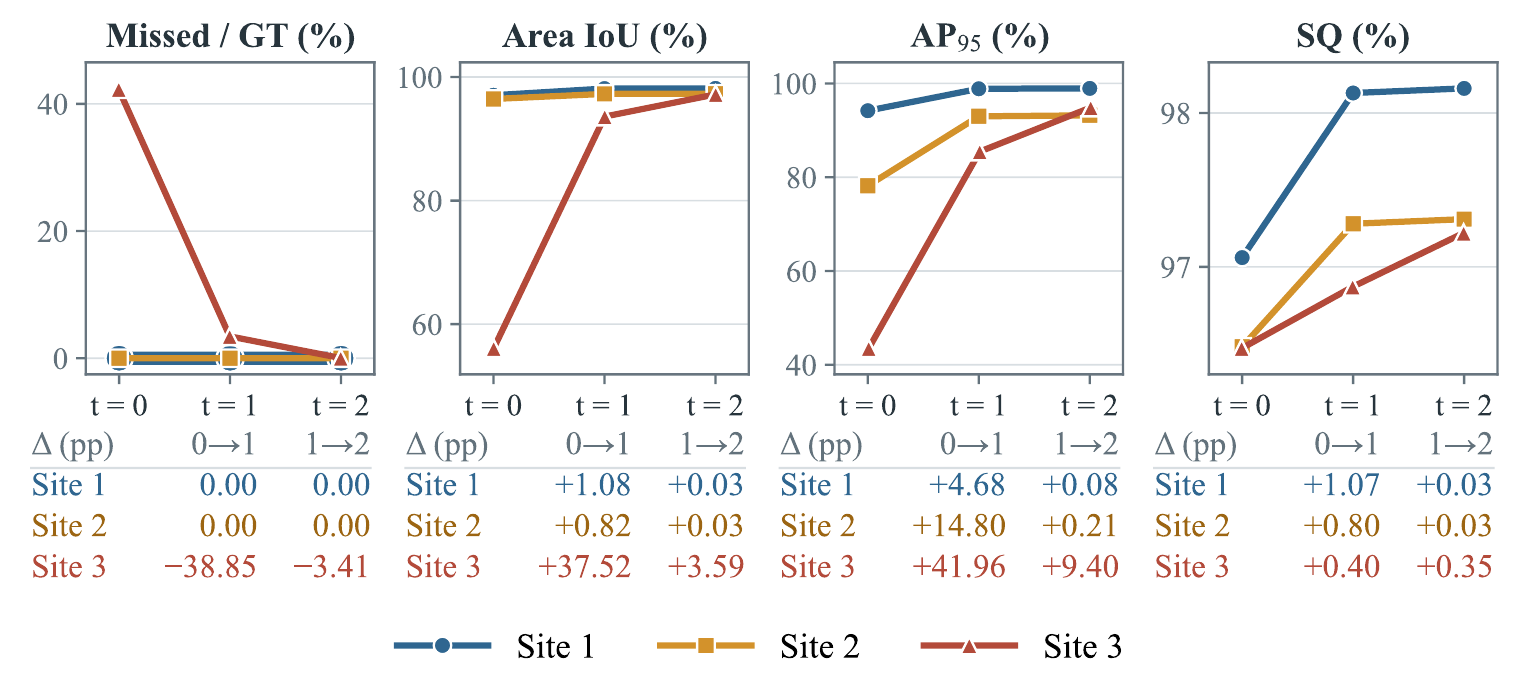}
    \caption{Per-site accuracy of MV-PVMapper at $t=0$, $1$ and $2$. Each panel shows one metric for the
    three sites; Missed/GT is the percentage of reference modules that are missed. The table under each
    panel gives the change in percentage points made by the first ($0\to1$) and second ($1\to2$)
    iteration. In the Missed/GT panel, the lines of Sites~1 and~2 coincide at zero.}
    \label{fig:iter_by_site}
\end{figure}

\end{evidencewithtext}
\FloatBarrier

\begin{evidencewithtext}
All omissions occur at Site~3, where the initial pass misses 42.29\% of the modules, mostly in the interiors of the densely packed arrays that cover the entire roofs of three buildings (Figure~\ref{fig:iteration}). Feedback reduces this share to 0.03\%, and the second iteration still adds 9.40 percentage points of $AP_{95}$ at this site.
\par

\begin{figure}[H]
    \centering
    \includegraphics[width=\textwidth]{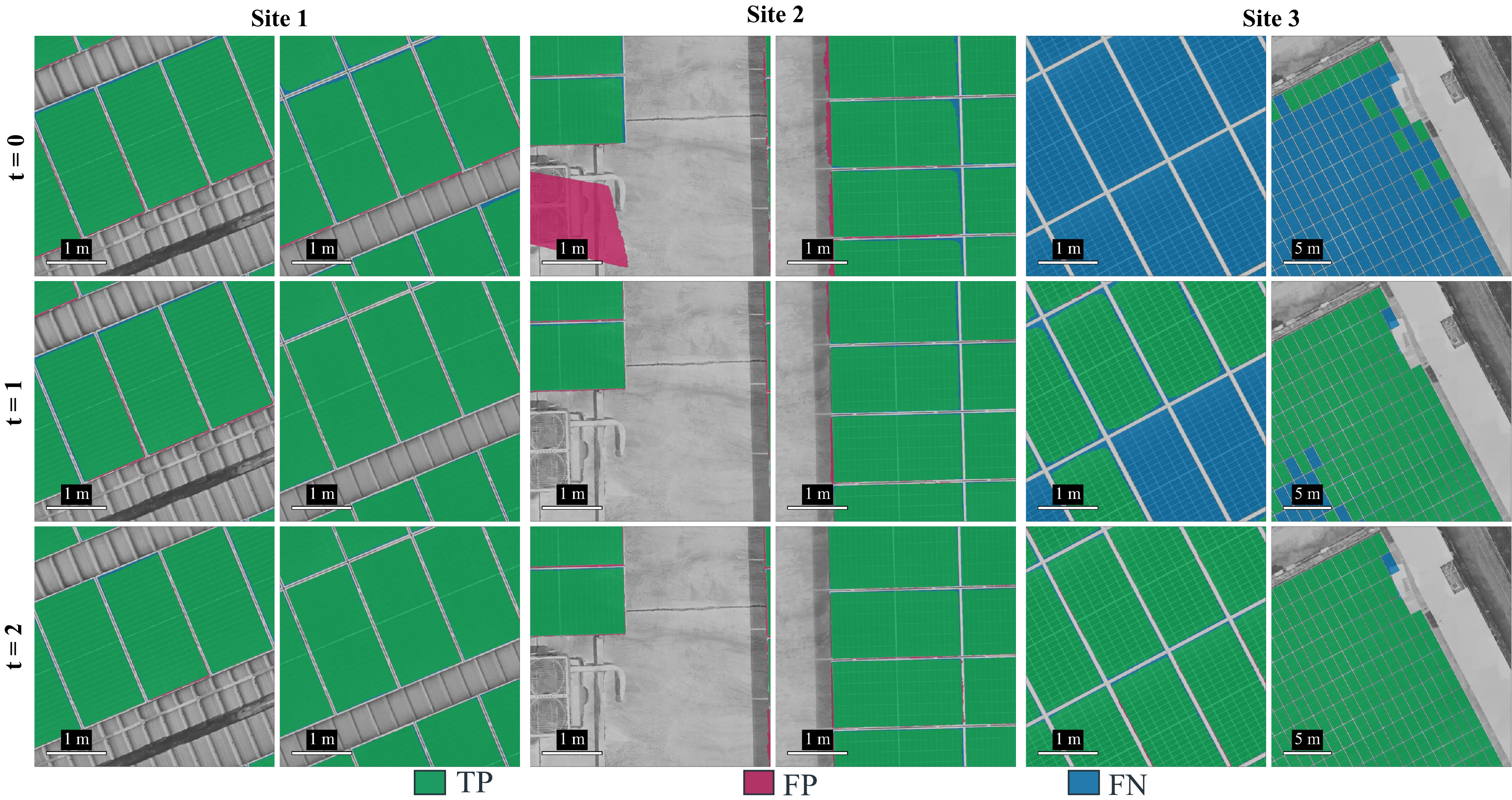}
    \caption{Module maps at $t=0$, $1$ and $2$, overlaid on the DOM of each site. Each site is shown by two
    crops; TP, FP and FN are shown at the pixel level.}
    \label{fig:iteration}
\end{figure}

\end{evidencewithtext}
\FloatBarrier

\begin{evidencewithtext}
Figure~\ref{fig:viewpoint_by_site} compares the nested viewpoint subsets at $t=0$, $1$ and $2$, showing how feedback changes the benefit of additional directions. At $t=0$, adding the first oblique direction raises area IoU above 94\% at Sites~1 and~2, but even all five directions leave 42.29\% of modules missed at Site~3. Feedback substantially reduces these omissions: at Site~3, nadir images alone reach an area IoU of 95.05\% at $t=2$, compared with 56.08\% for all five directions at $t=0$. The closer spacing of the $t=2$ results across viewpoint subsets, especially after the first oblique direction is added, indicates that feedback reduces the dependence on the number of input directions.
\par

\begin{figure}[H]
    \centering
    \includegraphics[width=\textwidth]{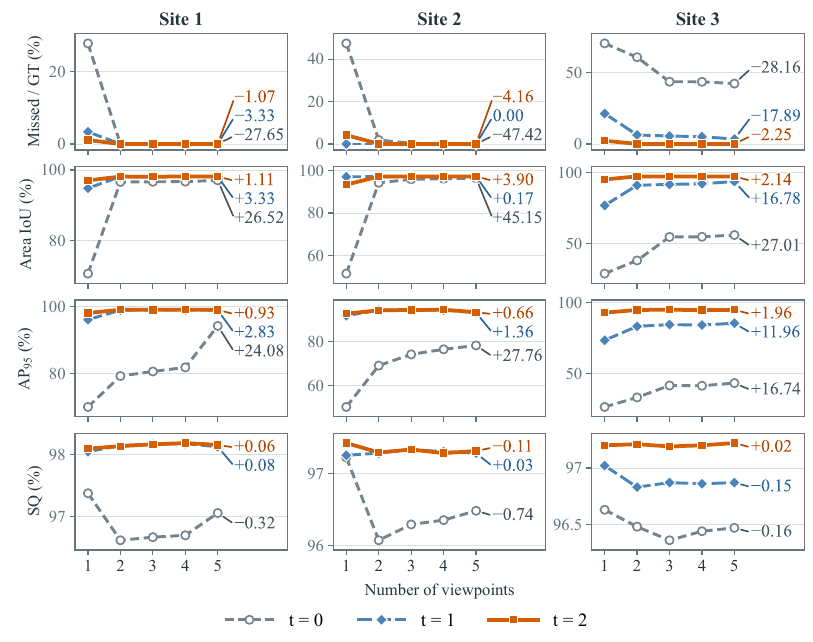}
    \caption{Per-site accuracy of the nested viewpoint subsets at $t=0$, $1$ and $2$. The horizontal axis
    gives the number of viewpoints: the nadir images alone, followed by one to four oblique directions
    added in a fixed order. The number at the end of each line is its change in percentage points from
    one to five viewpoints. Missed/GT is the percentage of reference modules that are missed.}
    \label{fig:viewpoint_by_site}
\end{figure}

\end{evidencewithtext}
\FloatBarrier

The benefit of a further feedback round nevertheless depends on the site and viewpoint configuration. At Site~3, the $t=1$ and $t=2$ curves remain clearly separated in missed-module rate and $AP_{95}$, whereas at Site~2 the nadir-only configuration misses no module at $t=1$ but 4.16\% at $t=2$. At $t=2$, increasing the number of directions produces only modest changes in mapping accuracy, particularly beyond the first oblique direction, with no consistent increase in $AP_{95}$. A possible explanation is that modules at these three sites are largely unoccluded in oblique views, so additional directions offer limited opportunities to reveal previously hidden module regions.
\par

\subsection{Implications, Applicability and Limitations}
\label{sec:limitations}

On the three evaluated rooftop plants, a frozen VFM combined with photogrammetric geometry and limited dimensional information produces a nearly complete module inventory with accurate boundaries. The geometry prior does not replace image interpretation; it helps select candidates at the scale of one physical module, while covering-view feedback allows reliable observations to improve other views. The resulting georeferenced instances provide the spatial objects needed to associate inspection findings with individual assets. Here, training-free means that no task-specific model training or weight updates are performed; it does not mean that the method requires no site information.

The input requirement also defines the scope of deployment. MV-PVMapper is designed for overlapping oblique UAV imagery with reconstructed camera poses and a DSM, rather than an isolated image or an orthophoto alone. Photogrammetric reconstruction is already used in UAV-based PV mapping and inspection workflows \citep{bommes2022georeferencing,cardoso2024automated}. Where such a survey is available, the method reuses the redundant source observations and reconstructed geometry; its additional site-specific input is a representative module size. The acquisition and reconstruction effort remains necessary where these data do not yet exist. Each feedback round processes every sliding window again, making the present implementation more suitable for offline map generation than real-time inspection.

The evaluated sites do not cover rooftops with highly irregular installations and several substantially different module specifications. A single representative size may then reject valid modules or admit fragments of larger ones. Lower image resolution, severe occlusion and larger installation tilts also require further evaluation.

We use SAM~3 as a representative model with the required prompting interface rather than compare a large set of segmentation backbones. The present experiments isolate the contribution of photogrammetric coordination while holding the backbone fixed. Near-complete recovery on these sites leaves limited scope for distinguishing models by instance count alone. Later or alternative promptable models may still improve boundary accuracy or reduce inference cost.


\section{Conclusion}
\label{sec:conclusion}

This study presents MV-PVMapper, a training-free framework for extracting georeferenced PV module instances from UAV multi-view images. It connects prompt-driven SAM~3 segmentation with object-space geometric verification and returns the fused modules to covering views as spatial prompts. Per-module slope correction transfers image contours through a locally coherent DSM surface, while the module-geometry score guides candidate selection, cross-view association and feedback. The resulting workflow links the complementary observations of a photogrammetric survey to a common module map without updating the segmentation-model weights.

On three rooftop plants containing 7\,711 reference modules, two feedback iterations produce 7\,710 matched instances with no false positives and one false negative. The framework achieves a macro-averaged AJI of 97.55\% and an $AP_{95}$ of 95.67\%, compared with 60.73\% and 60.51\% for DOM-IterSeg and 41.24\% and 27.52\% for MV-PixelVote. The ablations support the combined roles of covering-view feedback, local surface correction and geometric screening. These findings indicate that a pretrained image model can support module-level mapping when its predictions are assessed and coordinated using photogrammetric geometry.

\section*{CRediT authorship contribution statement}

\textbf{WenHao~Duan:} Conceptualization, Methodology, Software, Investigation, Validation, Visualization, Writing -- original draft.
\textbf{Qi~Chen:} Conceptualization, Methodology, Supervision, Project administration, Funding acquisition, Resources, Writing -- review \& editing.
\textbf{Penglei~Zhang:} Data curation, Formal analysis, Validation, Visualization.
\textbf{Ruizhe~Deng:} Methodology, Validation, Writing -- review \& editing.
\textbf{Zhiling~Guo:} Validation, Writing -- review \& editing.
\textbf{Haoran~Zhang:} Validation, Writing -- review \& editing.
\textbf{Yuntian~Chen:} Validation, Writing -- review \& editing.

\section*{Declaration of competing interest}

The authors declare that they have no known competing financial interests
or personal relationships that could have appeared to influence the work
reported in this paper.

\section*{Acknowledgements}

This work is supported by the National Natural Science Foundation of China under Grant~42371475, as well as by the High-Performance Computing Centers at Eastern Institute of Technology, Ningbo, and Ningbo Institute of Digital Twin.

\section*{Declaration of generative AI and AI-assisted technologies}

During the preparation of this work, the authors used GPT-6 Astra (OpenAI)
to improve the language and readability of the manuscript. After using
this tool, the authors reviewed and edited the content as needed and take
full responsibility for the content of the published article.

\bibliographystyle{elsarticle-harv}
\bibliography{references}



\end{document}